\PassOptionsToPackage{unicode}{hyperref}
\PassOptionsToPackage{hyphens}{url}
\PassOptionsToPackage{dvipsnames,svgnames,x11names}{xcolor}
\documentclass[
  twocolumn]{article}
\usepackage{xcolor}
\usepackage{amsmath,amssymb}
\usepackage{iftex}
\ifPDFTeX
  \usepackage[T1]{fontenc}
  \usepackage[utf8]{inputenc}
  \usepackage{textcomp} 
\else 
  \usepackage{unicode-math} 
  \defaultfontfeatures{Scale=MatchLowercase}
  \defaultfontfeatures[\rmfamily]{Ligatures=TeX,Scale=1}
\fi
\usepackage{lmodern}
\ifPDFTeX\else
\fi
\IfFileExists{upquote.sty}{\usepackage{upquote}}{}
\IfFileExists{microtype.sty}{
  \usepackage[]{microtype}
  \UseMicrotypeSet[protrusion]{basicmath} 
}{}
\makeatletter
\@ifundefined{KOMAClassName}{
  \IfFileExists{parskip.sty}{%
    \usepackage{parskip}
  }{
    \setlength{\parindent}{0pt}
    \setlength{\parskip}{6pt plus 2pt minus 1pt}}
}{
  \KOMAoptions{parskip=half}}
\makeatother
\makeatletter
\ifx\paragraph\undefined\else
  \let\oldparagraph\paragraph
  \renewcommand{\paragraph}{
    \@ifstar
      \xxxParagraphStar
      \xxxParagraphNoStar
  }
  \newcommand{\xxxParagraphStar}[1]{\oldparagraph*{#1}\mbox{}}
  \newcommand{\xxxParagraphNoStar}[1]{\oldparagraph{#1}\mbox{}}
\fi
\ifx\subparagraph\undefined\else
  \let\oldsubparagraph\subparagraph
  \renewcommand{\subparagraph}{
    \@ifstar
      \xxxSubParagraphStar
      \xxxSubParagraphNoStar
  }
  \newcommand{\xxxSubParagraphStar}[1]{\oldsubparagraph*{#1}\mbox{}}
  \newcommand{\xxxSubParagraphNoStar}[1]{\oldsubparagraph{#1}\mbox{}}
\fi
\makeatother

\usepackage{color}
\usepackage{fancyvrb}

\DefineVerbatimEnvironment{Highlighting}{Verbatim}{commandchars=\\\{\}}
\usepackage{framed}
\definecolor{shadecolor}{RGB}{241,243,245}
\newenvironment{Shaded}{\begin{snugshade}}{\end{snugshade}}

\newcommand{\BuiltInTok}[1]{\textcolor[rgb]{0.00,0.23,0.31}{#1}}

\newcommand{\CommentTok}[1]{\textcolor[rgb]{0.37,0.37,0.37}{#1}}

\newcommand{\ControlFlowTok}[1]{\textcolor[rgb]{0.00,0.23,0.31}{\textbf{#1}}}

\newcommand{\DecValTok}[1]{\textcolor[rgb]{0.68,0.00,0.00}{#1}}

\newcommand{\FloatTok}[1]{\textcolor[rgb]{0.68,0.00,0.00}{#1}}

\newcommand{\ImportTok}[1]{\textcolor[rgb]{0.00,0.46,0.62}{#1}}

\newcommand{\KeywordTok}[1]{\textcolor[rgb]{0.00,0.23,0.31}{\textbf{#1}}}
\newcommand{\NormalTok}[1]{\textcolor[rgb]{0.00,0.23,0.31}{#1}}
\newcommand{\OperatorTok}[1]{\textcolor[rgb]{0.37,0.37,0.37}{#1}}

\newcommand{\SpecialCharTok}[1]{\textcolor[rgb]{0.37,0.37,0.37}{#1}}
\newcommand{\SpecialStringTok}[1]{\textcolor[rgb]{0.13,0.47,0.30}{#1}}
\newcommand{\StringTok}[1]{\textcolor[rgb]{0.13,0.47,0.30}{#1}}
\newcommand{\VariableTok}[1]{\textcolor[rgb]{0.07,0.07,0.07}{#1}}

\usepackage{longtable,booktabs,array}
\usepackage{calc} 
\usepackage{etoolbox}
\makeatletter
\patchcmd\longtable{\par}{\if@noskipsec\mbox{}\fi\par}{}{}
\makeatother
\IfFileExists{footnotehyper.sty}{\usepackage{footnotehyper}}{\usepackage{footnote}}
\makesavenoteenv{longtable}
\usepackage{graphicx}
\makeatletter
\newsavebox\pandoc@box
\newcommand*\pandocbounded[1]{
  \sbox\pandoc@box{#1}%
  \Gscale@div\@tempa{\textheight}{\dimexpr\ht\pandoc@box+\dp\pandoc@box\relax}%
  \Gscale@div\@tempb{\linewidth}{\wd\pandoc@box}%
  \ifdim\@tempb\p@<\@tempa\p@\let\@tempa\@tempb\fi
  \ifdim\@tempa\p@<\p@\scalebox{\@tempa}{\usebox\pandoc@box}%
  \else\usebox{\pandoc@box}%
  \fi%
}
\def\fps@figure{htbp}
\makeatother

\NewDocumentCommand\citeproctext{}{}

\makeatletter
 \let\@cite@ofmt\@firstofone
 \def\@biblabel#1{}
 \def\@cite#1#2{{#1\if@tempswa , #2\fi}}
\makeatother
\newlength{\cslhangindent}
\newlength{\csllabelwidth}
\newenvironment{CSLReferences}[2] 
 {\begin{list}{}{%
  \setlength{\itemindent}{0pt}
  \setlength{\leftmargin}{0pt}
  \setlength{\parsep}{0pt}
  \ifodd #1
   \setlength{\leftmargin}{\cslhangindent}
   \setlength{\itemindent}{-1\cslhangindent}
  \fi
  \setlength{\itemsep}{#2\baselineskip}}}
 {\end{list}}
\usepackage{calc}

\providecommand{\tightlist}{%
  \setlength{\itemsep}{0pt}\setlength{\parskip}{0pt}}

\usepackage{arxiv}
\usepackage{orcidlink}
\usepackage{amsmath}
\usepackage[T1]{fontenc}
\usepackage{dirtree}
\DefineVerbatimEnvironment{Highlighting}{Verbatim}{commandchars=\\\{\},fontsize=\small}
\makeatletter
\@ifpackageloaded{caption}{}{\usepackage{caption}}
\AtBeginDocument{%
\ifdefined\contentsname
  \renewcommand*\contentsname{Table of contents}
\else
  \newcommand\contentsname{Table of contents}
\fi
\ifdefined\listfigurename
  \renewcommand*\listfigurename{List of Figures}
\else
  \newcommand\listfigurename{List of Figures}
\fi
\ifdefined\listtablename
  \renewcommand*\listtablename{List of Tables}
\else
  \newcommand\listtablename{List of Tables}
\fi
\ifdefined\figurename
  \renewcommand*\figurename{Figure}
\else
  \newcommand\figurename{Figure}
\fi
\ifdefined\tablename
  \renewcommand*\tablename{Table}
\else
  \newcommand\tablename{Table}
\fi
}
\@ifpackageloaded{float}{}{\usepackage{float}}
\floatstyle{ruled}
\@ifundefined{c@chapter}{\newfloat{codelisting}{h}{lop}}{\newfloat{codelisting}{h}{lop}[chapter]}
\floatname{codelisting}{Listing}

\makeatother
\makeatletter
\@ifpackageloaded{caption}{}{\usepackage{caption}}
\@ifpackageloaded{subcaption}{}{\usepackage{subcaption}}
\makeatother
\usepackage{bookmark}
\IfFileExists{xurl.sty}{\usepackage{xurl}}{} 
\makeatletter
\@ifundefined{xmpquote}{\newcommand{\xmpquote}[1]{#1}}{}
\makeatother
\hypersetup{
  pdftitle={Optimization with SpotOptim},
  pdfkeywords={\xmpquote{Surrogate modeling}, \xmpquote{Sequential
parameter optimization}, \xmpquote{Bayesian
optimization}, \xmpquote{Hyperparameter tuning}, \xmpquote{Kriging}},
  colorlinks=true,
  linkcolor={blue},
  filecolor={Maroon},
  citecolor={Blue},
  urlcolor={Blue},
  pdfcreator={LaTeX via pandoc}}

\newcommand{\runninghead}{A Preprint }
\renewcommand{\runninghead}{Working Paper }
\title{Optimization with SpotOptim}
\author{\textbf{Thomas
Bartz-Beielstein}~\orcidlink{0000-0002-5938-5158}\\\\Bartz \& Bartz
GmbH, 51643 Gummersbach,
Germany\\\\\href{mailto:bartzbeielstein@gmail.com}{bartzbeielstein@gmail.com}}
\date{}
\begin{document}
\maketitle
\begin{abstract}
The \texttt{spotoptim} package implements surrogate-model-based
optimization of expensive black-box functions in Python. Building on two
decades of Sequential Parameter Optimization (SPO) methodology, it
provides a Kriging-based optimization loop with Expected Improvement,
support for continuous, integer, and categorical variables, noise-aware
evaluation via Optimal Computing Budget Allocation (OCBA), and
multi-objective extensions. A success-rate-based restart mechanism
detects stagnation while preserving the best solution found. The package
returns scipy-compatible \texttt{OptimizeResult} objects and accepts any
scikit-learn-compatible surrogate model. Built-in TensorBoard logging
provides real-time monitoring of convergence and surrogate quality. This
report describes the architecture and module structure of spotoptim,
provides worked examples including neural network hyperparameter tuning,
and compares the framework with BoTorch, Optuna, Ray Tune, BOHB, SMAC,
and Hyperopt. The package is open-source (AGPL-3.0).
\end{abstract}
{\bfseries\itshape Keywords}
\def\sep{\textbullet\ }
Surrogate modeling \sep Sequential parameter optimization \sep Bayesian
optimization \sep Hyperparameter tuning \sep 
Kriging

\section{Introduction}\label{sec-introduction}

Problems in engineering, simulation, and machine and deep learning (or
generally in artificial intelligence) require the optimization of
functions that are expensive to evaluate. Training a deep neural network
to convergence, running a computational fluid dynamics simulation, or
evaluating a reinforcement learning policy may take minutes to hours per
function call, making exhaustive search impractical.
Surrogate-model-based optimization addresses this challenge by
constructing a cheap statistical approximation of the objective function
and using it to guide the search toward promising regions of the
parameter space (Forrester et al. 2008; Gramacy 2020). Sequential
Parameter Optimization (SPO) was introduced by Bartz-Beielstein et al.
(2005) as a principled framework for tuning the parameters of
metaheuristic algorithms. Rather than relying on default settings or
ad-hoc parameter sweeps, SPO fits a Kriging (Gaussian process) model to
the observed function evaluations, selects the next evaluation point by
optimizing an acquisition function such as Expected Improvement (EI)
(Jones et al. 1998), and iterates until the evaluation budget is
exhausted. This approach generalizes the Efficient Global Optimization
algorithm (Jones et al. 1998) to a broader class of tuning and
optimization problems, including noisy objectives and mixed variable
types.

The SPO methodology has been implemented in several software packages
over the past two decades. The original R package SPOT, which was
available on the Comprehensive R Archive Network (CRAN)\footnote{\url{https://cran.r-project.org/web/packages/SPOT/index.html}},
provided the first publicly available implementation and was used
extensively in the companion volume ``Hyperparameter Tuning for Machine
and Deep Learning with R'' (Bartz et al. 2022)\footnote{With more than
  150k accesses, it is one of the most popular publications in the
  field. See
  \url{https://link.springer.com/book/10.1007/978-981-19-5170-1}.}. An
overview of the SPOT methodology and its R implementation is given by
Bartz-Beielstein et al. (2021). The R package was subsequently ported to
Python as SpotPython, which extended the framework with PyTorch
integration and a hyperparameter tuning cookbook (Bartz-Beielstein
2023a). The \texttt{spotoptim} package\footnote{\url{https://gitlab.git.nrw/thk-f10/spotsevenlab/spotoptim}}
is the current generation of this lineage. It is a complete rewrite that
preserves the core SPO algorithm while modernizing the architecture,
improving extensibility, and integrating with the Python scientific
computing ecosystem. The package is part of a family of related tools.
Together, these packages form an ecosystem for optimization-driven
scientific computing research and practice.

The contributions of this report are threefold. First, it positions
spotoptim within the landscape of hyperparameter optimization frameworks
by comparing it with BoTorch, Optuna, Ray Tune, BOHB, SMAC, and Hyperopt
(Section~\ref{sec-related}). Second, it provides a comprehensive
description of the \texttt{spotoptim} architecture, covering the
optimization algorithm, surrogate models, acquisition functions, and
supporting modules (Section~\ref{sec-algorithm} through
Section~\ref{sec-modules}). Third, it presents worked examples that
demonstrate the package API for tasks ranging from simple function
optimization to end-to-end neural network hyperparameter tuning
(Section~\ref{sec-examples} and Section~\ref{sec-hpt}).

The remainder of this report is organized as follows.
Section~\ref{sec-related} reviews related work and competing frameworks.
Section~\ref{sec-examples} introduces the package through three
progressively complex examples. Section~\ref{sec-algorithm} describes
the SPO algorithm as implemented in \texttt{spotoptim}.
Section~\ref{sec-modules} details each module of the package.
Section~\ref{sec-hpt} presents an end-to-end hyperparameter tuning
workflow. Section~\ref{sec-outlook} concludes with a summary.

\section{Related Work}\label{sec-related}

Hyperparameter optimization has received sustained attention over the
past decade, resulting in several mature software frameworks. These
tools differ along multiple axes: the search strategy they employ
(random, bandit-based, or model-based), the type of surrogate model they
use (if any), their parallelism model (single-machine or distributed),
and the interface they present to the user. This section reviews the
most widely used frameworks and highlights how SPO, as implemented in
\texttt{spotoptim}, relates to each of them.

Hyperopt (Bergstra et al. 2011) introduced Tree-structured Parzen
Estimators (TPE) as an alternative to Gaussian-process-based Bayesian
optimization. TPE avoids the \(\mathcal{O}(n^3)\) cost of fitting a
Gaussian process to \(n\) observations, making it more scalable as
evaluations accumulate. However, it does not yield a global surrogate
model and therefore cannot produce uncertainty estimates or support
acquisition functions like EI in their standard form.

Optuna (Akiba et al. 2019) is a popular hyperparameter optimization
framework in the Python ecosystem. It employs a ``define-by-run'' API in
which the search space is specified implicitly through trial
suggestions, rather than declared upfront. The default search strategy
uses TPE. Optuna also supports Covariance Matrix Adaptation Evolution
Strategy (CMA-ES) and provides a pruning mechanism based on successive
halving that allows unpromising trials to be terminated early.

Bayesian Optimization and Hyperband (BOHB) (Falkner et al. 2018)
combines Bayesian optimization with Hyperband, a multi-fidelity method
that allocates resources adaptively across trials. The Bayesian
component uses TPE as its surrogate, similar to Optuna. BOHB's key
contribution is the integration of early stopping into the
surrogate-based search, allowing it to discard poorly performing
configurations after partial training. This multi-fidelity approach is
effective when intermediate performance measures (such as validation
loss after a few epochs) are available. In contrast, \texttt{spotoptim}
treats the objective function as a black box that returns a single
scalar per evaluation and does not currently incorporate multi-fidelity
scheduling.

Sequential Model-based Algorithm Configuration (SMAC) (Hutter et al.
2011) is the framework most closely related to SPO in its algorithmic
philosophy. Like SPO, SMAC iteratively fits a surrogate model and
selects new configurations by optimizing an acquisition function. The
key difference lies in the choice of surrogate: SMAC uses random forests
which handle high-dimensional and categorical parameter spaces well but
do not provide the smooth, differentiable uncertainty estimates that
Gaussian processes offer. SMAC has its roots in SPO (Hutter et al.
2010): similar to SPO, it was originally designed for algorithm
configuration, where the goal is to find parameter settings that
minimize the runtime or solution quality of a target algorithm across a
distribution of problem instances. \texttt{spotoptim} targets a broader
class of optimization problems, including engineering design and
simulation-based optimization, and returns scipy-compatible results that
integrate directly with the scientific Python ecosystem.

Ray Tune (Liaw et al. 2018) is a distributed hyperparameter tuning
platform built on top of the Ray framework. Rather than implementing a
single search strategy, Ray Tune serves as an orchestrator that wraps
external search algorithms including Optuna, Hyperopt, and Bayesian
optimization libraries. Its primary strength lies in scalable trial
scheduling across clusters, making it well-suited for large-scale
distributed training. While Ray Tune excels at distributed scheduling,
it is not itself a surrogate-based optimizer and delegates the actual
search logic to external backends.

BoTorch (Balandat et al. 2020) is a PyTorch-based library for Bayesian
optimization developed at Meta. It provides Gaussian process surrogates
and enables efficient handling of batch, multi-objective, and
constrained settings. BoTorch is designed as a modular research toolkit
and assumes familiarity with PyTorch idioms such as tensors, devices,
and custom training loops. In contrast, \texttt{spotoptim} targets
practitioners working within the scipy/scikit-learn ecosystem.

Several features distinguish SPO and its implementation in
\texttt{spotoptim} from the frameworks reviewed above. First,
\texttt{spotoptim} uses Kriging as its default surrogate, providing
principled uncertainty quantification through the predictive variance of
the Gaussian process. This enables acquisition functions such as EI
(Jones et al. 1998) and Probability of Improvement with a sound
statistical foundation. Second, the package returns scipy-compatible
\texttt{OptimizeResult} objects, allowing results to be consumed by any
tool in the scipy ecosystem without conversion. Third,
\texttt{spotoptim} natively supports mixed variable types (continuous,
integer, and categorical) with appropriate handling within the surrogate
model. Fourth, noisy objectives are handled through built-in repeated
evaluations combined with Optimal Computing Budget Allocation (OCBA)
(Chen 1996; Chen et al. 2000), a feature not available in any of the
competing frameworks reviewed here. Fifth, multi-objective optimization
is supported and scalarization via desirability functions is available
(Bartz-Beielstein 2025a, 2025b). Finally, the surrogate interface
follows the scikit-learn estimator convention
(\texttt{fit}/\texttt{predict}), so Kriging can be replaced by any
compatible surrogate model (Section~\ref{sec-surrogate}).

\section{Simple Examples}\label{sec-examples}

This section introduces the \texttt{spotoptim} API through three
progressively complex examples. Each example is self-contained and
demonstrates a different aspect of the optimization workflow.

\subsection{Minimizing the Sphere
Function}\label{minimizing-the-sphere-function}

The simplest use case is the optimization of a scalar-valued function
over continuous variables. The following code minimizes the sphere
function \(f(\mathbf{x}) = \sum_{i=1}^d x_i^2\), where \(d\) denotes the
number of dimensions, in two dimensions:

\begin{Shaded}
\begin{Highlighting}[]
\ImportTok{from}\NormalTok{ spotoptim }\ImportTok{import}\NormalTok{ SpotOptim}
\ImportTok{from}\NormalTok{ spotoptim.function }\ImportTok{import}\NormalTok{ sphere}

\NormalTok{opt }\OperatorTok{=}\NormalTok{ SpotOptim(}
\NormalTok{    fun}\OperatorTok{=}\NormalTok{sphere,}
\NormalTok{    bounds}\OperatorTok{=}\NormalTok{[(}\OperatorTok{{-}}\DecValTok{5}\NormalTok{, }\DecValTok{5}\NormalTok{), (}\OperatorTok{{-}}\DecValTok{5}\NormalTok{, }\DecValTok{5}\NormalTok{)],}
\NormalTok{    max\_iter}\OperatorTok{=}\DecValTok{20}\NormalTok{,}
\NormalTok{    n\_initial}\OperatorTok{=}\DecValTok{10}\NormalTok{,}
\NormalTok{    seed}\OperatorTok{=}\DecValTok{0}\NormalTok{,}
\NormalTok{)}
\NormalTok{result }\OperatorTok{=}\NormalTok{ opt.optimize()}
\BuiltInTok{print}\NormalTok{(}\SpecialStringTok{f"Best value: }\SpecialCharTok{\{}\NormalTok{result}\SpecialCharTok{.}\NormalTok{fun}\SpecialCharTok{:.6f\}}\SpecialStringTok{"}\NormalTok{)}
\BuiltInTok{print}\NormalTok{(}\SpecialStringTok{f"Best point: }\SpecialCharTok{\{}\NormalTok{result}\SpecialCharTok{.}\NormalTok{x}\SpecialCharTok{\}}\SpecialStringTok{"}\NormalTok{)}
\end{Highlighting}
\end{Shaded}

\begin{verbatim}
Best value: 0.000001
Best point: [-0.00016718  0.00071419]
\end{verbatim}

Three ingredients are required: a callable \texttt{fun} that accepts an
\((n, d)\) array (where \(n\) is the number of samples to evaluate) and
returns an \((n,)\) array, a list of \texttt{bounds} as
\texttt{(lower,\ upper)} tuples, and an evaluation budget via
\texttt{max\_iter}. The \texttt{n\_initial} parameter controls how many
points are evaluated in the initial Latin Hypercube design before the
surrogate-based sequential phase begins. The \texttt{optimize()} method
returns a \texttt{scipy.optimize.OptimizeResult}, which carries the best
point (\texttt{result.x}), the corresponding objective value
(\texttt{result.fun}), and the total number of function evaluations
(\texttt{result.nfev}), among other fields.

\subsection{Expected Improvement with Explicit
Kriging}\label{expected-improvement-with-explicit-kriging}

The default acquisition function is \texttt{"y"} (predicted value),
which performs pure exploitation by selecting the point where the
surrogate predicts the lowest value. For problems with multiple local
minima, EI provides a better exploration-exploitation trade-off, because
it accounts for both the predicted value and the surrogate's
uncertainty:

\begin{equation}\protect\phantomsection\label{eq-ei}{
\begin{aligned}
\text{EI}(\mathbf{x}) &= (y_{\min} - \mu(\mathbf{x})) \, \Phi(Z) + \sigma(\mathbf{x}) \, \phi(Z), \\
Z &= \frac{y_{\min} - \mu(\mathbf{x})}{\sigma(\mathbf{x})}
\end{aligned}
}\end{equation}

where \(\mu(\mathbf{x})\) and \(\sigma(\mathbf{x})\) are the Kriging
mean and standard deviation, \(y_{\min}\) is the best observed value,
and \(\Phi\) and \(\phi\) are the standard normal cumulative
distribution function and probability density function, respectively
(Forrester et al. 2008).

\begin{Shaded}
\begin{Highlighting}[]
\ImportTok{from}\NormalTok{ spotoptim }\ImportTok{import}\NormalTok{ SpotOptim}
\ImportTok{from}\NormalTok{ spotoptim.surrogate }\ImportTok{import}\NormalTok{ Kriging}
\ImportTok{from}\NormalTok{ spotoptim.function }\ImportTok{import}\NormalTok{ rosenbrock}

\NormalTok{kriging }\OperatorTok{=}\NormalTok{ Kriging(}
\NormalTok{    method}\OperatorTok{=}\StringTok{"regression"}\NormalTok{,}
\NormalTok{    noise}\OperatorTok{=}\FloatTok{1e{-}3}\NormalTok{, seed}\OperatorTok{=}\DecValTok{0}\NormalTok{,}
\NormalTok{)}

\NormalTok{opt }\OperatorTok{=}\NormalTok{ SpotOptim(}
\NormalTok{    fun}\OperatorTok{=}\NormalTok{rosenbrock,}
\NormalTok{    bounds}\OperatorTok{=}\NormalTok{[(}\OperatorTok{{-}}\DecValTok{2}\NormalTok{, }\DecValTok{2}\NormalTok{), (}\OperatorTok{{-}}\DecValTok{2}\NormalTok{, }\DecValTok{2}\NormalTok{)],}
\NormalTok{    surrogate}\OperatorTok{=}\NormalTok{kriging,}
\NormalTok{    acquisition}\OperatorTok{=}\StringTok{"ei"}\NormalTok{,}
\NormalTok{    max\_iter}\OperatorTok{=}\DecValTok{25}\NormalTok{,}
\NormalTok{    n\_initial}\OperatorTok{=}\DecValTok{10}\NormalTok{,}
\NormalTok{    seed}\OperatorTok{=}\DecValTok{0}\NormalTok{,}
\NormalTok{)}
\NormalTok{result }\OperatorTok{=}\NormalTok{ opt.optimize()}

\BuiltInTok{print}\NormalTok{(}\SpecialStringTok{f"Best value: }\SpecialCharTok{\{}\NormalTok{result}\SpecialCharTok{.}\NormalTok{fun}\SpecialCharTok{:.6f\}}\SpecialStringTok{"}\NormalTok{)}
\BuiltInTok{print}\NormalTok{(}\SpecialStringTok{f"Best point: }\SpecialCharTok{\{}\NormalTok{result}\SpecialCharTok{.}\NormalTok{x}\SpecialCharTok{\}}\SpecialStringTok{"}\NormalTok{)}
\end{Highlighting}
\end{Shaded}

\begin{verbatim}
Best value: 0.013070
Best point: [0.89033462 0.7894651 ]
\end{verbatim}

Here the Kriging surrogate is constructed explicitly with a noise term
for regularization. The \texttt{acquisition="ei"} argument switches the
infill criterion from predicted value to EI. Any surrogate model that
supports \texttt{predict(X,\ return\_std=True)} can be used with EI and
with Probability of Improvement, which is available via the
\texttt{acquisition="pi"} argument (see Section~\ref{sec-optimizer}).

\subsection{Mixed Variable Types}\label{mixed-variable-types}

Many practical optimization problems involve a mixture of continuous,
integer, and categorical variables. \texttt{spotoptim} handles this
natively through the \texttt{var\_type} parameter:

\begin{Shaded}
\begin{Highlighting}[]
\ImportTok{import}\NormalTok{ numpy }\ImportTok{as}\NormalTok{ np}
\ImportTok{from}\NormalTok{ spotoptim }\ImportTok{import}\NormalTok{ SpotOptim}

\KeywordTok{def}\NormalTok{ mixed\_objective(X):}
\NormalTok{    X }\OperatorTok{=}\NormalTok{ np.atleast\_2d(X)}
\NormalTok{    continuous }\OperatorTok{=}\NormalTok{ X[:, }\DecValTok{0}\NormalTok{]}
\NormalTok{    integer\_val }\OperatorTok{=}\NormalTok{ X[:, }\DecValTok{1}\NormalTok{]}
\NormalTok{    factor\_val }\OperatorTok{=}\NormalTok{ X[:, }\DecValTok{2}\NormalTok{]}
    \ControlFlowTok{return}\NormalTok{ (continuous}\OperatorTok{**}\DecValTok{2}
            \OperatorTok{+}\NormalTok{ (integer\_val }\OperatorTok{{-}} \DecValTok{3}\NormalTok{)}\OperatorTok{**}\DecValTok{2}
            \OperatorTok{+}\NormalTok{ factor\_val)}

\NormalTok{opt }\OperatorTok{=}\NormalTok{ SpotOptim(}
\NormalTok{    fun}\OperatorTok{=}\NormalTok{mixed\_objective,}
\NormalTok{    bounds}\OperatorTok{=}\NormalTok{[(}\OperatorTok{{-}}\FloatTok{5.0}\NormalTok{, }\FloatTok{5.0}\NormalTok{), (}\DecValTok{0}\NormalTok{, }\DecValTok{10}\NormalTok{), (}\DecValTok{0}\NormalTok{, }\DecValTok{4}\NormalTok{)],}
\NormalTok{    var\_type}\OperatorTok{=}\NormalTok{[}\StringTok{"float"}\NormalTok{, }\StringTok{"int"}\NormalTok{, }\StringTok{"factor"}\NormalTok{],}
\NormalTok{    var\_name}\OperatorTok{=}\NormalTok{[}\StringTok{"x\_cont"}\NormalTok{, }\StringTok{"x\_int"}\NormalTok{, }\StringTok{"x\_cat"}\NormalTok{],}
\NormalTok{    max\_iter}\OperatorTok{=}\DecValTok{25}\NormalTok{,}
\NormalTok{    n\_initial}\OperatorTok{=}\DecValTok{10}\NormalTok{,}
\NormalTok{    seed}\OperatorTok{=}\DecValTok{0}\NormalTok{,}
\NormalTok{)}
\NormalTok{result }\OperatorTok{=}\NormalTok{ opt.optimize()}

\BuiltInTok{print}\NormalTok{(}\SpecialStringTok{f"Best value: }\SpecialCharTok{\{}\NormalTok{result}\SpecialCharTok{.}\NormalTok{fun}\SpecialCharTok{:.6f\}}\SpecialStringTok{"}\NormalTok{)}
\BuiltInTok{print}\NormalTok{(}\SpecialStringTok{f"Best point: }\SpecialCharTok{\{}\NormalTok{result}\SpecialCharTok{.}\NormalTok{x}\SpecialCharTok{\}}\SpecialStringTok{"}\NormalTok{)}
\end{Highlighting}
\end{Shaded}

\begin{verbatim}
Best value: 0.000001
Best point: [-9.98879117e-04  3.00000000e+00  0.00000000e+00]
\end{verbatim}

The three supported variable types are \texttt{"float"} (continuous),
\texttt{"int"} (integer-constrained, rounded after surrogate
prediction), and \texttt{"factor"} (categorical, encoded internally).
When \texttt{var\_type} is omitted, all variables default to
\texttt{"float"}.

\section{The SPO Algorithm}\label{sec-algorithm}

The default optimization loop implemented in
\texttt{SpotOptim.optimize()} follows the general structure of
surrogate-model-based optimization, also known as Bayesian optimization
when the surrogate is a Gaussian process (Gramacy 2020). The algorithm
proceeds in two phases: an initial design phase that builds a
preliminary picture of the response surface, and a sequential phase that
iteratively refines the surrogate model and proposes new evaluation
points.

In the initial design phase, \texttt{n\_initial} points are generated
according to a space-filling design. The default is a quasi-Monte Carlo
Latin Hypercube Sampling (LHS) design (QMC-LHS), which ensures that the
marginal distribution of each variable is well-covered. Alternative
designs include Sobol sequences, regular grids, uniform random sampling,
and clustered designs. The user may also provide a custom initial design
via the \texttt{X0} argument. All initial points are evaluated on the
true objective function, and the results form the initial training set
for the surrogate. In the sequential phase, the algorithm repeats the
following steps until the evaluation budget (\texttt{max\_iter}) or the
wall-clock time limit (\texttt{max\_time}) is reached:

\begin{enumerate}
\def\labelenumi{\arabic{enumi}.}
\tightlist
\item
  Fit the surrogate model to all observed data \((X, \mathbf{y})\).
\item
  Optimize the acquisition function over the search space to identify
  the next candidate point \(\mathbf{x}_{\text{new}}\).
\item
  Evaluate \(f(\mathbf{x}_{\text{new}})\) on the true objective.
\item
  Append the new observation to the data set and update running
  statistics.
\end{enumerate}

Three acquisition functions are supported, which are optimized over the
search space using a configurable acquisition optimizer
(\texttt{differential\_evolution} by default). \texttt{spotoptim} runs a
strictly sequential loop: after each evaluation the new observation is
appended to the data set and the surrogate is refitted before the next
infill point is proposed, so every decision uses all information
gathered so far. By default a single infill point is proposed per
iteration. Setting \texttt{n\_infill\_points} greater than one proposes
several candidates from the same surrogate fit; the candidates are
evaluated one after another and the surrogate is refitted once the batch
has been processed, which reduces the number of surrogate refits at the
cost of acting on a slightly older surrogate. The following example
proposes two infill points per surrogate update:

\begin{Shaded}
\begin{Highlighting}[]
\ImportTok{from}\NormalTok{ spotoptim }\ImportTok{import}\NormalTok{ SpotOptim}
\ImportTok{from}\NormalTok{ spotoptim.function }\ImportTok{import}\NormalTok{ sphere}

\NormalTok{opt }\OperatorTok{=}\NormalTok{ SpotOptim(}
\NormalTok{    fun}\OperatorTok{=}\NormalTok{sphere,}
\NormalTok{    bounds}\OperatorTok{=}\NormalTok{[(}\OperatorTok{{-}}\DecValTok{5}\NormalTok{, }\DecValTok{5}\NormalTok{), (}\OperatorTok{{-}}\DecValTok{5}\NormalTok{, }\DecValTok{5}\NormalTok{)],}
\NormalTok{    max\_iter}\OperatorTok{=}\DecValTok{50}\NormalTok{,}
\NormalTok{    n\_initial}\OperatorTok{=}\DecValTok{10}\NormalTok{,}
\NormalTok{    seed}\OperatorTok{=}\DecValTok{0}\NormalTok{,}
\NormalTok{    n\_infill\_points}\OperatorTok{=}\DecValTok{2}\NormalTok{,  }\CommentTok{\# candidates per surrogate update}
\NormalTok{)}
\NormalTok{result }\OperatorTok{=}\NormalTok{ opt.optimize()}

\BuiltInTok{print}\NormalTok{(}\SpecialStringTok{f"Best value: }\SpecialCharTok{\{}\NormalTok{result}\SpecialCharTok{.}\NormalTok{fun}\SpecialCharTok{:.6f\}}\SpecialStringTok{"}\NormalTok{)}
\BuiltInTok{print}\NormalTok{(}\SpecialStringTok{f"Total evaluations: }\SpecialCharTok{\{}\NormalTok{result}\SpecialCharTok{.}\NormalTok{nfev}\SpecialCharTok{\}}\SpecialStringTok{"}\NormalTok{)}
\end{Highlighting}
\end{Shaded}

\begin{verbatim}
Best value: 0.000000
Total evaluations: 50
\end{verbatim}

For noisy objective functions, \texttt{spotoptim} supports repeated
evaluations at each design point. The surrogate is fitted on the mean
values across repeats, reducing the influence of noise. When the noise
level varies across the search space, OCBA can be enabled through the
\texttt{ocba\_delta} parameter. OCBA allocates additional evaluation
budget to the most promising and most uncertain designs. The method was
introduced by Chen (1996), and the asymptotically optimal allocation
rule was established by Chen et al. (2000; see also Chen 2010). Its
integration into the SPO framework was developed by Bartz-Beielstein and
Friese (2011) and Bartz-Beielstein et al. (2011). This combination of
repeated evaluations and adaptive budget allocation provides a
principled approach to noisy optimization that is unique among the
frameworks discussed in Section~\ref{sec-related}.

When the optimizer stalls, automatic restarts can help escape local
minima. \texttt{spotoptim} tracks a rolling success rate that measures
the fraction of recent evaluations that improved upon the incumbent best
value. A sliding window of size \texttt{window\_size} records whether
each sequential evaluation achieved a new best; the success rate is the
number of successes divided by the window length. By default
\texttt{window\_size} is set to \texttt{restart\_after\_n} (or 100 if
\texttt{restart\_after\_n} is also unset), so the success rate reflects
performance over the full restart horizon. When no improvement has
occurred for a full window, the success rate drops to zero, signaling
stagnation. The \texttt{restart\_after\_n} parameter (default 100)
specifies how many consecutive iterations with a zero success rate must
elapse before a restart is triggered. Upon restart, the optimizer
generates a fresh initial design and re-initializes the surrogate. If
\texttt{restart\_inject\_best} is \texttt{True} (the default), the best
solution found so far is injected into the new initial design,
preserving accumulated knowledge while allowing the surrogate to explore
a different region of the search space. The following example shows how
to configure the success-rate-based restart mechanism:

\begin{Shaded}
\begin{Highlighting}[]
\ImportTok{from}\NormalTok{ spotoptim }\ImportTok{import}\NormalTok{ SpotOptim}
\ImportTok{from}\NormalTok{ spotoptim.function }\ImportTok{import}\NormalTok{ sphere}

\NormalTok{opt }\OperatorTok{=}\NormalTok{ SpotOptim(}
\NormalTok{    fun}\OperatorTok{=}\NormalTok{sphere,}
\NormalTok{    bounds}\OperatorTok{=}\NormalTok{[(}\OperatorTok{{-}}\DecValTok{5}\NormalTok{, }\DecValTok{5}\NormalTok{), (}\OperatorTok{{-}}\DecValTok{5}\NormalTok{, }\DecValTok{5}\NormalTok{)],}
\NormalTok{    max\_iter}\OperatorTok{=}\DecValTok{20}\NormalTok{,}
\NormalTok{    n\_initial}\OperatorTok{=}\DecValTok{10}\NormalTok{,}
\NormalTok{    seed}\OperatorTok{=}\DecValTok{42}\NormalTok{,}
\NormalTok{    window\_size}\OperatorTok{=}\DecValTok{5}\NormalTok{,}
\NormalTok{    restart\_after\_n}\OperatorTok{=}\DecValTok{10}\NormalTok{,}
\NormalTok{    restart\_inject\_best}\OperatorTok{=}\VariableTok{True}\NormalTok{,}
\NormalTok{    verbose}\OperatorTok{=}\VariableTok{False}\NormalTok{,}
\NormalTok{)}
\NormalTok{result }\OperatorTok{=}\NormalTok{ opt.optimize()}

\BuiltInTok{print}\NormalTok{(}\SpecialStringTok{f"Success rate: }\SpecialCharTok{\{}\NormalTok{opt}\SpecialCharTok{.}\NormalTok{success\_rate}\SpecialCharTok{:.2f\}}\SpecialStringTok{"}\NormalTok{)}
\BuiltInTok{print}\NormalTok{(}\SpecialStringTok{f"Best value: }\SpecialCharTok{\{}\NormalTok{result}\SpecialCharTok{.}\NormalTok{fun}\SpecialCharTok{:.6f\}}\SpecialStringTok{"}\NormalTok{)}
\BuiltInTok{print}\NormalTok{(}\SpecialStringTok{f"Evaluations: }\SpecialCharTok{\{}\NormalTok{result}\SpecialCharTok{.}\NormalTok{nfev}\SpecialCharTok{\}}\SpecialStringTok{"}\NormalTok{)}
\end{Highlighting}
\end{Shaded}

\begin{verbatim}
Success rate: 0.40
Best value: 0.000000
Evaluations: 20
\end{verbatim}

A small \texttt{window\_size} makes the success rate sensitive to short
bursts of improvement, while a larger window smooths out isolated lucky
evaluations. A low \texttt{restart\_after\_n} triggers frequent
restarts, which favors exploration over exploitation. A high value
allows the optimizer to persist longer in a region before restarting.
The success rate is also available programmatically via the
\texttt{success\_rate} attribute, enabling custom termination logic or
logging.

Restarts alone cannot detect the situation in which the optimizer has
exhausted the information content of the search space: repeated restarts
may keep resampling similar regions without ever improving on the
incumbent. To save evaluation budget in this situation,
\texttt{spotoptim} implements a patience-based early-stopping rule
through the \texttt{max\_restarts} parameter. When set to a non-negative
integer \(N\), the outer loop terminates after \(N\) consecutive
restarts that fail to improve the best objective value. This rule is
analogous to the \texttt{no\_progress\_loss} helper of Hyperopt
(Bergstra et al. 2011), the \texttt{ExperimentPlateauStopper} of Ray
Tune, and the \texttt{terminate\_cost\_threshold} scenario in SMAC
(Hutter et al. 2011). The resulting \texttt{OptimizeResult} has
\texttt{success=True} and a message of the form ``Optimization early
stopped: no improvement for \(N\) consecutive restarts'', which lets
downstream pipelines distinguish a graceful plateau termination from a
budget exhaustion.

\begin{Shaded}
\begin{Highlighting}[]
\ImportTok{from}\NormalTok{ spotoptim }\ImportTok{import}\NormalTok{ SpotOptim}
\ImportTok{from}\NormalTok{ spotoptim.function }\ImportTok{import}\NormalTok{ sphere}

\NormalTok{opt }\OperatorTok{=}\NormalTok{ SpotOptim(}
\NormalTok{    fun}\OperatorTok{=}\NormalTok{sphere,}
\NormalTok{    bounds}\OperatorTok{=}\NormalTok{[(}\OperatorTok{{-}}\DecValTok{5}\NormalTok{, }\DecValTok{5}\NormalTok{), (}\OperatorTok{{-}}\DecValTok{5}\NormalTok{, }\DecValTok{5}\NormalTok{)],}
\NormalTok{    max\_iter}\OperatorTok{=}\DecValTok{200}\NormalTok{,}
\NormalTok{    n\_initial}\OperatorTok{=}\DecValTok{5}\NormalTok{,}
\NormalTok{    restart\_after\_n}\OperatorTok{=}\DecValTok{3}\NormalTok{,}
\NormalTok{    window\_size}\OperatorTok{=}\DecValTok{3}\NormalTok{,}
\NormalTok{    max\_restarts}\OperatorTok{=}\DecValTok{2}\NormalTok{,}
\NormalTok{    seed}\OperatorTok{=}\DecValTok{0}\NormalTok{,}
\NormalTok{    verbose}\OperatorTok{=}\VariableTok{False}\NormalTok{,}
\NormalTok{)}
\NormalTok{result }\OperatorTok{=}\NormalTok{ opt.optimize()}

\BuiltInTok{print}\NormalTok{(result.message.splitlines()[}\DecValTok{0}\NormalTok{])}
\BuiltInTok{print}\NormalTok{(}\SpecialStringTok{f"Evaluations used: }\SpecialCharTok{\{}\NormalTok{result}\SpecialCharTok{.}\NormalTok{nfev}\SpecialCharTok{\}}\SpecialStringTok{"}\NormalTok{)}
\end{Highlighting}
\end{Shaded}

\begin{verbatim}
Optimization early stopped: no improvement for 2 consecutive restarts
Evaluations used: 20
\end{verbatim}

The role of each parameter in the example is summarized in
Table~\ref{tbl-early-stopping-params}.

\begin{table}

\caption{\label{tbl-early-stopping-params}Role of each parameter in the early-stopping example.}

\centering{

\centering
\small
\begin{tabular}{@{}lp{0.55\linewidth}@{}}
\hline
Parameter & Role \\
\hline
\texttt{max\_iter=200} & Hard evaluation budget. Stops the run if nothing else triggers first. \\
\texttt{n\_initial=5} & Each LHS design (initial \emph{and} every restart) uses 5 points. \\
\texttt{window\_size=3} & Success rate is averaged over the last 3 infill iterations. \\
\texttt{restart\_after\_n=3} & 3 consecutive iterations with success-rate = 0 trigger a restart. \\
\texttt{max\_restarts=2} & After 2 restarts in a row that fail to improve \texttt{best\_y\_}, stop early. \\
\hline
\end{tabular}

}

\end{table}%

The resulting execution trace on the \texttt{sphere} objective can be
summarized as follows. The initial design (evaluations 1-5) draws an LHS
sample in \([-5, 5]^2\). During the subsequent infill phase, the
surrogate is refit on every iteration, and early infills typically
improve the incumbent because \texttt{sphere} is smooth; the rate of
improvement slows as the search approaches the minimum at \((0, 0)\).
Once three infills in a row fail to improve the incumbent (the
success-rate window is zero for three consecutive iterations), the first
restart is triggered: the current best is injected, and a fresh
five-point LHS design is drawn. Because the injected best is already
near zero, neither the new LHS nor the next three infills can beat it,
so \texttt{\_restarts\_without\_improvement} is incremented to 1. Since
\(1 < 2\) the rule does not yet fire, and a second restart starts. Its
outcome is the same, incrementing the counter to 2. The check
\(2 \geq 2\) now succeeds and the outer loop breaks before a third
restart is executed. The typical evaluation count is therefore
approximately \(5 + 3 + 5 + 3 + 5 + 3 = 24\) evaluations, well below the
\texttt{max\_iter=200} budget. The returned \texttt{OptimizeResult} has
\texttt{success\ =\ True} and a message starting with ``Optimization
early stopped: no improvement for 2 consecutive restarts'';
\texttt{opt.restarts\_results\_} has length 3 (initial run plus two
restarts) and \texttt{opt.\_early\_stopped} is \texttt{True}.

Several boundary cases follow directly from this logic. Setting
\texttt{max\_restarts=None} (the default) disables the rule and
preserves the pre-existing behavior in which the optimizer runs until
\texttt{max\_iter} or \texttt{max\_time} is reached. Raising
\texttt{restart\_after\_n} so that the first restart cannot be triggered
within the evaluation budget has the same practical effect. If the
objective is one on which restarts can genuinely improve the incumbent,
the counter is reset on every productive restart and the rule may never
fire. Conversely, \texttt{max\_restarts=0} represents zero tolerance:
the run stops after the very first non-improving restart --- on
\texttt{sphere}, this is usually immediately after the first restart
cycle. A future release will generalize this single-purpose rule into a
pluggable stopping-criterion framework that additionally supports
absolute target values, expected-improvement thresholds, and
user-supplied callbacks, covering the full spectrum of stopping patterns
discussed in Section~\ref{sec-related}.

\section{Modules}\label{sec-modules}

The \texttt{spotoptim} codebase is organized into focused modules
(subpackages), each responsible for a specific aspect of the
optimization workflow. Figure \ref{fig-dirtree} shows the top-level
directory structure. This section describes each module, its purpose,
and its key components. Key abbreviations used in the figure and
throughout this section include multi-layer perceptron (MLP) and
principal component analysis (PCA).

The base installation, obtained with \texttt{pip\ install\ spotoptim}
and requiring Python 3.13 or later, is deliberately lean: it depends
only on NumPy, SciPy, scikit-learn, pandas, tabulate, and dill, which
suffices for the core Kriging optimization loop, the analytical test
functions, and the reporting utilities. Capabilities that rely on
heavier dependencies are provided as optional extras. The \texttt{torch}
extra adds the neural network surrogates, the PyTorch objective and
dataset utilities of the \texttt{nn} and \texttt{data} subpackages, and
TensorBoard logging; the \texttt{viz} extra adds the Matplotlib- and
seaborn-based figures of the \texttt{plot}, \texttt{inspection},
\texttt{eda}, and multi-objective modules; the \texttt{stats} extra adds
statsmodels-based statistical utilities; and the \texttt{remote} extra
adds requests-based remote objective evaluation. The bundle
\texttt{spotoptim{[}all{]}} installs all four, and the neural network
and visualization examples in this report assume at least the
\texttt{torch} and \texttt{viz} extras. Consistent with this lean
design, only \texttt{SpotOptim}, \texttt{Kriging},
\texttt{SimpleKriging}, and a few core helpers are re-exported from the
top-level \texttt{spotoptim} namespace; all other components are
imported from their respective subpackages, as the examples below
illustrate.

\begin{figure*}[ht]
\dirtree{%
.1 src/spotoptim/.
.2 SpotOptim.py\DTcomment{Core optimizer}.
.2 core/\DTcomment{Protocol, storage, experiment control}.
.2 optimizer/\DTcomment{Acquisition, schedule-free optimizer, scipy wrapper}.
.2 surrogate/\DTcomment{Kriging, MLP surrogate, Nystroem}.
.2 nn/\DTcomment{PyTorch MLP, LinearRegressor}.
.2 function/\DTcomment{Objective functions (single-/multi-objective, remote, torch)}.
.2 sampling/\DTcomment{LHS, Sobol, grid, clustered designs}.
.2 reporting/\DTcomment{Results extraction, analysis utilities}.
.2 plot/\DTcomment{Surrogate visualization, contour, multi-objective plots}.
.2 utils/\DTcomment{Boundaries, transforms, PCA, OCBA, TensorBoard}.
.2 mo/\DTcomment{Multi-objective: Morris--Mitchell, Pareto front}.
.2 hyperparameters/\DTcomment{Parameter set management for neural network tuning}.
.2 data/\DTcomment{Dataset loaders (e.g., DiabetesDataset)}.
.2 datasets/\DTcomment{Bundled CSV datasets}.
.2 inspection/\DTcomment{Model/surrogate inspection}.
.2 factor\_analyzer/\DTcomment{Factor analysis}.
.2 eda/\DTcomment{Exploratory data analysis}.
.2 tricands/\DTcomment{Triangulation-based candidate generation}.
}
\caption{Top-level directory structure of the \texttt{spotoptim} package.}\label{fig-dirtree}
\end{figure*}

\subsection{The SpotOptim Class}\label{sec-spotoptim-class}

The \texttt{SpotOptim} class in \texttt{spotoptim.SpotOptim} is the
central orchestrator. Its constructor accepts the objective function,
bounds, and a comprehensive set of configuration parameters that control
every aspect of the optimization: the surrogate model, acquisition
function and optimizer, variable types and transformations, evaluation
budget, noise handling, and restart policy. All parameters are stored in
a \texttt{SpotOptimConfig} dataclass and can be accessed as attributes
of the optimizer instance. The most commonly used constructor parameters
are \texttt{fun} (the objective function), \texttt{bounds} (a list of
lower/upper tuples), \texttt{max\_iter} (total evaluation budget
including the initial design), \texttt{n\_initial} (number of initial
design points), \texttt{surrogate} (default:
\texttt{Kriging(method="regression")}), \texttt{acquisition}
(\texttt{"y"}, \texttt{"ei"}, or \texttt{"pi"}), \texttt{var\_type}
(list of \texttt{"float"}, \texttt{"int"}, \texttt{"factor"}), and
\texttt{seed} (for reproducibility). The \texttt{optimize()} method
executes the algorithm described in Section~\ref{sec-algorithm} and
returns a \texttt{scipy.optimize.OptimizeResult} with fields \texttt{x}
(best point), \texttt{fun} (best objective value), \texttt{nfev} (total
evaluations), \texttt{nit} (sequential iterations), \texttt{success},
and \texttt{message}. The full evaluated data are available as
\texttt{result.X} and \texttt{result.y}, allowing post-hoc analysis
without re-running the optimization.

Variable transformations can be applied through the \texttt{var\_trans}
parameter. For example, \texttt{var\_trans={[}"log10",\ None{]}}
optimizes the first variable in \(\log_{10}\) space internally while
specifying bounds in natural scale, which is useful for parameters that
span several orders of magnitude such as learning rates.

\subsection{Core Infrastructure}\label{sec-core}

The \texttt{core} subpackage provides foundational components.
\texttt{SpotOptimProtocol} (defined in \texttt{core/protocol.py}) is a
structural typing protocol (PEP 544) that declares the interface
extracted modules expect from the optimizer. Modules such as
\texttt{optimizer.acquisition} and \texttt{reporting.analysis} accept
any object matching this protocol rather than importing the concrete
\texttt{SpotOptim} class, avoiding circular imports and facilitating
independent testing. The \texttt{core.storage} module manages the
optimizer's internal data arrays through functions like
\texttt{init\_storage()} and \texttt{update\_storage()}, which handle
appending new evaluation results, updating running statistics, and
tracking the best solution found so far. \texttt{ExperimentControl} is a
dataclass that bundles dataset, model class, hyperparameters, device
settings, and training parameters into a single object for PyTorch-based
experiment workflows.

\subsection{Surrogate Models}\label{sec-surrogate}

The \texttt{surrogate} subpackage contains three surrogate
implementations. \texttt{Kriging} is the default and models the
objective as a Gaussian process with a Gaussian (squared-exponential)
kernel, yielding both a mean prediction \(\mu(\mathbf{x})\) and a
standard deviation \(\sigma(\mathbf{x})\) that is essential for
uncertainty-aware acquisition functions. Its key parameters include
\texttt{method} (see below), \texttt{noise} (regularization term),
\texttt{min\_theta} and \texttt{max\_theta} (bounds for log-scaled
kernel hyperparameters), and \texttt{seed}; a call to
\texttt{predict(X,\ return\_std=True)} returns both outputs.

The kernel hyperparameters \(\boldsymbol{\theta}\) are estimated by
maximizing the concentrated log-likelihood using differential evolution.
Following Forrester et al. (2008),\footnote{Specifically, Section 2.4
  ``Kriging'' for the core predictor and likelihood, and Section 6
  ``Surrogate Modeling of Noisy Data'' for the \texttt{"regression"} and
  \texttt{"reinterpolation"} methods. The Python code is based on
  \texttt{likelihood.m} (concentrated log-likelihood) and
  \texttt{pred.m} (prediction and error estimation) from the book's
  codebase.} three fitting modes are available via the \texttt{method}
argument: \texttt{"regression"} (default) fits a generalized
least-squares model, \texttt{"interpolation"} passes exactly through the
data points, and \texttt{"reinterpolation"} applies Forrester's
correction for noisy data. The implementation is validated against the
Matlab code of Forrester et al. (2008).

The Kriging implementation in SPO uses flexible kernel functions that
extend naturally to non-continuous parameter spaces. For categorical and
combinatorial variables, appropriate distance or similarity measures
replace the standard Euclidean distance in the correlation function,
enabling the surrogate to model landscapes over discrete, permutation,
or mixed search spaces (Bartz-Beielstein and Zaefferer 2017; Zaefferer
and Bartz-Beielstein 2016). This line of research has produced kernels
for permutation-based problems using tailored distance measures with
automated selection via maximum likelihood estimation (Zaefferer, Stork,
and Bartz-Beielstein 2014; Zaefferer, Stork, Friese, et al. 2014), as
well as kernels for hierarchical and conditional parameter spaces
arising in algorithm configuration (Gentile et al. 2021, 2018).

\begin{Shaded}
\begin{Highlighting}[]
\ImportTok{import}\NormalTok{ numpy }\ImportTok{as}\NormalTok{ np}
\ImportTok{from}\NormalTok{ spotoptim.surrogate }\ImportTok{import}\NormalTok{ Kriging}

\NormalTok{X\_train }\OperatorTok{=}\NormalTok{ np.array([[}\FloatTok{0.0}\NormalTok{], [}\FloatTok{1.0}\NormalTok{], [}\FloatTok{3.0}\NormalTok{], [}\FloatTok{4.0}\NormalTok{]])}
\NormalTok{y\_train }\OperatorTok{=}\NormalTok{ np.array([}\FloatTok{0.0}\NormalTok{, }\FloatTok{1.0}\NormalTok{, }\FloatTok{9.0}\NormalTok{, }\FloatTok{16.0}\NormalTok{])}

\NormalTok{model }\OperatorTok{=}\NormalTok{ Kriging(method}\OperatorTok{=}\StringTok{"regression"}\NormalTok{, seed}\OperatorTok{=}\DecValTok{0}\NormalTok{)}
\NormalTok{model.fit(X\_train, y\_train)}

\NormalTok{X\_test }\OperatorTok{=}\NormalTok{ np.array([[}\FloatTok{0.5}\NormalTok{], [}\FloatTok{2.0}\NormalTok{], [}\FloatTok{3.5}\NormalTok{]])}
\NormalTok{y\_pred, y\_std }\OperatorTok{=}\NormalTok{ model.predict(}
\NormalTok{    X\_test, return\_std}\OperatorTok{=}\VariableTok{True}
\NormalTok{)}
\end{Highlighting}
\end{Shaded}

\texttt{SimpleKriging} is a lightweight alternative for simple
continuous problems where computational speed takes priority over
flexibility. \texttt{MLPSurrogate} uses an MLP, which is useful when the
response surface is highly non-linear or when the number of data points
exceeds the practical limits of Kriging's \(\mathcal{O}(n^3)\) fitting
cost. Alternatively, a Nystroem approximation module
(\texttt{surrogate/nystroem.py}) provides further scalability for large
datasets. Uncertainty estimates from \texttt{MLPSurrogate} are obtained
by performing multiple forward passes with dropout enabled and computing
the empirical variance across passes.

The surrogate interface follows the scikit-learn estimator convention.
Any model that implements \texttt{fit(X,\ y)} and \texttt{predict(X)}
can be passed as the \texttt{surrogate} argument to \texttt{SpotOptim}.
For acquisition functions that require uncertainty (\texttt{"ei"},
\texttt{"pi"}), the model should additionally support
\texttt{predict(X,\ return\_std=True)}. This makes it straightforward to
use scikit-learn's \texttt{GaussianProcessRegressor} with custom
kernels, or any other regression model, as a drop-in replacement for
Kriging.

Beyond single-surrogate optimization, \texttt{spotoptim} supports
multi-surrogate scheduling. The \texttt{surrogate} parameter accepts a
list of surrogate models together with a \texttt{prob\_surrogate} vector
that specifies the selection probability for each model. At every
surrogate refit step, one model is drawn at random according to these
weights and used for the next acquisition cycle. This introduces
diversity into the search: different surrogate types may fit different
regions of the landscape better, and alternating between them can reduce
the risk of systematic model bias. Each surrogate can also have its own
\texttt{max\_surrogate\_points} budget, passed as a list of the same
length. If \texttt{prob\_surrogate} is omitted, uniform weights are
assigned automatically. The following example combines a Kriging model
(selected with probability 0.7) and a random forest (selected with
probability 0.3):

\begin{Shaded}
\begin{Highlighting}[]
\ImportTok{from}\NormalTok{ spotoptim }\ImportTok{import}\NormalTok{ SpotOptim}
\ImportTok{from}\NormalTok{ spotoptim.surrogate }\ImportTok{import}\NormalTok{ Kriging}
\ImportTok{from}\NormalTok{ sklearn.ensemble }\ImportTok{import}\NormalTok{ RandomForestRegressor}
\ImportTok{from}\NormalTok{ spotoptim.function }\ImportTok{import}\NormalTok{ sphere}

\NormalTok{kriging }\OperatorTok{=}\NormalTok{ Kriging(method}\OperatorTok{=}\StringTok{"regression"}\NormalTok{, seed}\OperatorTok{=}\DecValTok{0}\NormalTok{)}
\NormalTok{rf }\OperatorTok{=}\NormalTok{ RandomForestRegressor(}
\NormalTok{    n\_estimators}\OperatorTok{=}\DecValTok{50}\NormalTok{, random\_state}\OperatorTok{=}\DecValTok{0}
\NormalTok{)}

\NormalTok{opt }\OperatorTok{=}\NormalTok{ SpotOptim(}
\NormalTok{    fun}\OperatorTok{=}\NormalTok{sphere,}
\NormalTok{    bounds}\OperatorTok{=}\NormalTok{[(}\OperatorTok{{-}}\DecValTok{5}\NormalTok{, }\DecValTok{5}\NormalTok{), (}\OperatorTok{{-}}\DecValTok{5}\NormalTok{, }\DecValTok{5}\NormalTok{)],}
\NormalTok{    max\_iter}\OperatorTok{=}\DecValTok{30}\NormalTok{,}
\NormalTok{    n\_initial}\OperatorTok{=}\DecValTok{10}\NormalTok{,}
\NormalTok{    seed}\OperatorTok{=}\DecValTok{0}\NormalTok{,}
\NormalTok{    surrogate}\OperatorTok{=}\NormalTok{[kriging, rf],}
\NormalTok{    prob\_surrogate}\OperatorTok{=}\NormalTok{[}\FloatTok{0.7}\NormalTok{, }\FloatTok{0.3}\NormalTok{],}
\NormalTok{    max\_surrogate\_points}\OperatorTok{=}\NormalTok{[}\VariableTok{None}\NormalTok{, }\DecValTok{50}\NormalTok{],}
    \CommentTok{\# no cap for Kriging, 50 for RF}
\NormalTok{)}
\NormalTok{result }\OperatorTok{=}\NormalTok{ opt.optimize()}

\BuiltInTok{print}\NormalTok{(}\SpecialStringTok{f"Best value: }\SpecialCharTok{\{}\NormalTok{result}\SpecialCharTok{.}\NormalTok{fun}\SpecialCharTok{:.6f\}}\SpecialStringTok{"}\NormalTok{)}
\end{Highlighting}
\end{Shaded}

\begin{verbatim}
Best value: 0.000000
\end{verbatim}

\subsection{Acquisition and Infill}\label{sec-optimizer}

The \texttt{optimizer} subpackage implements the acquisition functions,
their optimizers, and infill-point selection.

The \texttt{acquisition} parameter selects which criterion is used to
propose the next evaluation point. Three options are available.
\emph{Predicted value} (\texttt{"y"}) selects the point where the
surrogate predicts the lowest (or highest, for maximization) objective
value. This is the simplest strategy and amounts to pure exploitation of
the current model. It is computationally cheap but can become trapped in
local minima because it does not account for surrogate uncertainty.
\emph{Expected Improvement} (\texttt{"ei"}) balances exploitation and
exploration by weighting the predicted improvement over the current best
value \(y_{\min}\) against the surrogate's predictive uncertainty
\(\sigma(\mathbf{x})\). The EI formula (Equation~\ref{eq-ei}) was
introduced in Section~\ref{sec-examples}. Points with high predicted
quality \emph{or} high uncertainty receive large EI values, which
encourages the optimizer to explore under-sampled regions.
\emph{Probability of Improvement} (\texttt{"pi"}) selects the point with
the highest probability of producing an objective value below the
current best \(y_{\min}\). Probability of Improvement tends to be more
exploitative than EI, because it only measures the probability of any
improvement, not its expected magnitude.

The \texttt{acquisition\_optimizer} parameter determines how the
acquisition function is maximized over the search space. Differential
evolution (the default) performs a global search and is robust across a
wide range of problem structures (Storn 1996). The triangulation
candidates approach of Gramacy et al. (2022) generates candidate points
geometrically from the Delaunay triangulation of existing evaluations,
producing both interior candidates at simplex centroids and fringe
candidates that extend toward the search space boundary (see also
Section~\ref{sec-tricands}). The hybrid \texttt{de\_tricands} mode,
which is still experimental and has not been analyzed so far, alternates
between the two methods with probability controlled by
\texttt{prob\_de\_tricands}. Standard scipy minimizers are also
supported for local refinement.

Multiple infill points per iteration can be requested via
\texttt{n\_infill\_points}, as described in Section~\ref{sec-algorithm}.
When the acquisition optimizer fails to find a valid new point (for
example due to a flat surrogate surface), a random fallback point is
generated within bounds. For problems with many evaluation points, the
\texttt{max\_surrogate\_points} parameter limits the number of data
points used for surrogate fitting, keeping computational cost manageable
as the number of evaluations grows. Points are selected using K-means
clustering with either a space-filling criterion (\texttt{"distant"}) or
a quality-based criterion (\texttt{"best"}).

\subsection{Neural Network Models}\label{sec-nn}

The \texttt{nn} subpackage provides two PyTorch modules designed for use
as objective functions and surrogates in hyperparameter tuning
workflows. The \texttt{MLP} class is a \texttt{torch.nn.Sequential}
subclass with configurable width, depth, activation, and dropout. The
architecture can be specified either explicitly through a
\texttt{hidden\_channels} list or compactly through \texttt{l1} (neurons
per hidden layer) and \texttt{num\_hidden\_layers}, which is the
representation used during hyperparameter tuning.

\begin{Shaded}
\begin{Highlighting}[]
\ImportTok{import}\NormalTok{ torch}
\ImportTok{from}\NormalTok{ spotoptim.nn }\ImportTok{import}\NormalTok{ MLP}

\NormalTok{model }\OperatorTok{=}\NormalTok{ MLP(}
\NormalTok{    in\_channels}\OperatorTok{=}\DecValTok{10}\NormalTok{,}
\NormalTok{    l1}\OperatorTok{=}\DecValTok{64}\NormalTok{,}
\NormalTok{    num\_hidden\_layers}\OperatorTok{=}\DecValTok{2}\NormalTok{,}
\NormalTok{    output\_dim}\OperatorTok{=}\DecValTok{1}\NormalTok{,}
\NormalTok{    dropout}\OperatorTok{=}\FloatTok{0.1}\NormalTok{,}
\NormalTok{)}
\end{Highlighting}
\end{Shaded}

\texttt{LinearRegressor} is a \texttt{torch.nn.Module} for regression
tasks that ranges from pure linear regression (with
\texttt{num\_hidden\_layers=0}) to a deep network with configurable
activation functions. Both classes provide a
\texttt{get\_default\_parameters()} class method that returns a
\texttt{ParameterSet} with sensible bounds for hyperparameter tuning,
and a \texttt{get\_optimizer()} method that maps string names to
\texttt{torch.optim} optimizer classes. Beyond standard PyTorch
optimizers, \texttt{spotoptim} bundles \texttt{AdamWScheduleFree}, a
schedule-free variant of AdamW that does not require a learning-rate
scheduler.

\subsection{Built-in Test Functions}\label{sec-functions}

The \texttt{function} subpackage contains analytical test functions for
benchmarking and testing. All functions accept a 2-D array of shape
\((n, d)\) and return a 1-D array of shape \((n,)\) for single-objective
functions, or \((n, m)\) for multi-objective functions, where \(m\)
denotes the number of objectives.

The single-objective functions include sphere, noisy sphere (sphere with
additive Gaussian noise), Rosenbrock (narrow curved valley, minimum at
\(\mathbf{1}\)), Ackley (multi-modal with many local minima), and
Michalewicz (steep valleys with a tunable steepness parameter).
Engineering benchmark functions include \texttt{wingwt} (wing weight
estimation, 9--10 dimensions from Forrester et al. (2008)),
\texttt{robot\_arm\_hard} (10-link robot arm maze navigation), and
\texttt{lennard\_jones} (atomic cluster potential, 39 dimensions for 13
atoms). Multi-objective functions include the ZDT family (\texttt{zdt1}
through \texttt{zdt6}), DTLZ problems (\texttt{dtlz1}, \texttt{dtlz2}),
Fonseca-Fleming, Schaffer N1, and Kursawe. Custom objective functions
can be defined by the user following the same array convention.

\subsection{Sampling and Experimental Designs}\label{sec-sampling}

The \texttt{sampling.design} module provides space-filling designs for
the initial evaluation phase. The default quasi-Monte Carlo Latin
Hypercube design (\texttt{generate\_qmc\_lhs\_design}) ensures that each
variable's marginal distribution is uniformly covered. Sobol sequences
(\texttt{generate\_sobol\_design}) provide quasi-random low-discrepancy
coverage that is particularly effective in higher dimensions. Regular
grids (\texttt{generate\_grid\_design}) place points at equal intervals;
the actual number of grid points is
\(\lfloor n_\text{design}^{1/d} \rfloor^d\), where \(n_\text{design}\)
is the requested number of points. Uniform random sampling
(\texttt{generate\_uniform\_design}) serves as a baseline, and clustered
designs (\texttt{generate\_clustered\_design}) produce non-uniform
distributions for testing optimizer robustness and generating so-called
``ill-conditioned'' designs.

\begin{Shaded}
\begin{Highlighting}[]
\ImportTok{from}\NormalTok{ spotoptim.sampling.design }\ImportTok{import}\NormalTok{ (}
\NormalTok{    generate\_qmc\_lhs\_design,}
\NormalTok{    generate\_sobol\_design,}
\NormalTok{)}

\NormalTok{bounds }\OperatorTok{=}\NormalTok{ [(}\OperatorTok{{-}}\DecValTok{5}\NormalTok{, }\DecValTok{5}\NormalTok{), (}\OperatorTok{{-}}\DecValTok{5}\NormalTok{, }\DecValTok{5}\NormalTok{)]}
\NormalTok{X\_lhs }\OperatorTok{=}\NormalTok{ generate\_qmc\_lhs\_design(}
\NormalTok{    bounds, n\_design}\OperatorTok{=}\DecValTok{20}\NormalTok{, seed}\OperatorTok{=}\DecValTok{0}
\NormalTok{)}
\NormalTok{X\_sobol }\OperatorTok{=}\NormalTok{ generate\_sobol\_design(}
\NormalTok{    bounds, n\_design}\OperatorTok{=}\DecValTok{32}\NormalTok{, seed}\OperatorTok{=}\DecValTok{0}
\NormalTok{)}
\end{Highlighting}
\end{Shaded}

A pre-computed initial design can be passed to
\texttt{SpotOptim.optimize()} via the \texttt{X0} parameter, allowing
the user to incorporate prior knowledge or to resume an optimization
from a previous set of evaluations.

\subsection{Reporting and Analysis}\label{sec-reporting}

The \texttt{reporting} subpackage extracts and formats optimization
results. \texttt{print\_best} displays the best parameter vector and
objective value in a human-readable format, with factor variables mapped
back to their string labels. \texttt{get\_results\_table} produces a
formatted table showing each variable's name, type, bounds, default
value, and tuned (best) value, with an optional importance score column.
\texttt{get\_design\_table} summarizes the search space before
optimization, listing variable types, bounds, and transformations. For
post-hoc analysis, \texttt{get\_importance} computes a correlation-based
importance score for each variable on a 0--100 scale, and
\texttt{sensitivity\_spearman} reports Spearman rank correlations
between each parameter and the objective value, together with p-values
and significance stars. These tools help identify which hyperparameters
have the strongest influence on performance, guiding subsequent
refinements to the search space.

\subsection{Visualization}\label{sec-plotting}

\begin{figure}[H]

\centering{

\pandocbounded{\includegraphics[keepaspectratio]{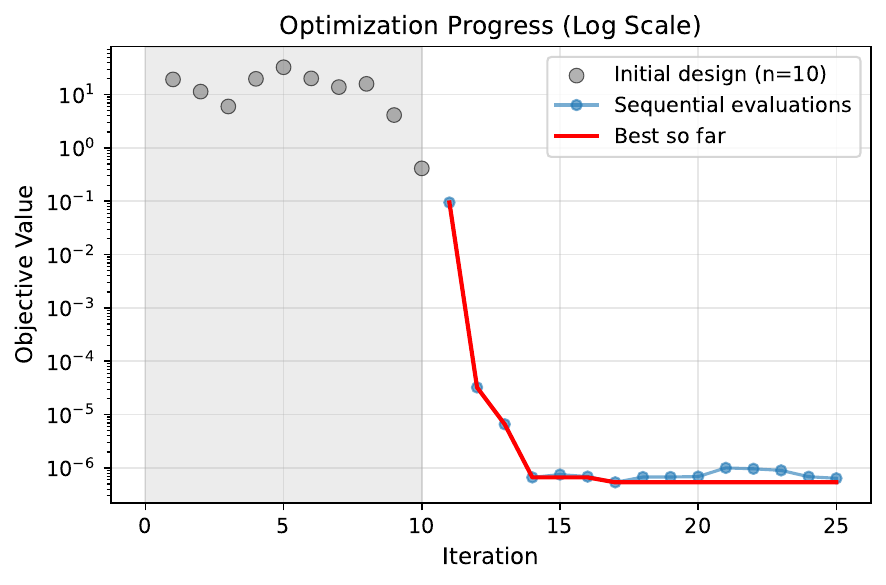}}

}

\caption{\label{fig-progress}Optimization progress for the sphere
function. Grey dots mark the initial Latin Hypercube design; subsequent
evaluations are connected by a line. The red curve shows the best
objective value found so far.}

\end{figure}

The \texttt{plot} subpackage provides several visualization functions.
\texttt{plot\_progress} displays the full evaluation history as a
scatter plot with a best-so-far curve overlaid, distinguishing initial
design points from sequential evaluations. \texttt{plot\_surrogate}
renders a 2x2 panel showing the fitted surrogate model for a selected
pair of dimensions: the top row contains 3-D surfaces of predictions and
prediction uncertainty, while the bottom row shows the corresponding
contour plots with evaluated points overlaid. \texttt{simple\_contour}
draws a quick 2-D filled contour of any callable over a rectangular
region, and \texttt{plot\_design\_points} creates a scatter plot of
evaluated points with hidden-dimension aggregation. Multi-objective
visualization is provided through \texttt{mo\_pareto\_optx\_plot}, which
shows Pareto-optimal points in the input space, and
\texttt{mo\_xy\_contour} and \texttt{mo\_xy\_surface} for
surrogate-based objective-space visualization. The following examples
use the sphere function optimized over \([-5, 5]^2\) with 25 iterations.
Figure~\ref{fig-progress} shows a typical progress plot. The initial
design points appear as grey dots in a shaded background region.
Sequential evaluations are connected by a line, and the red curve traces
the best objective value found so far.

\begin{Shaded}
\begin{Highlighting}[]
\ImportTok{from}\NormalTok{ spotoptim }\ImportTok{import}\NormalTok{ SpotOptim}
\ImportTok{from}\NormalTok{ spotoptim.function }\ImportTok{import}\NormalTok{ sphere}
\ImportTok{from}\NormalTok{ spotoptim.plot.visualization }\ImportTok{import}\NormalTok{ (}
\NormalTok{    plot\_progress}
\NormalTok{)}

\NormalTok{opt }\OperatorTok{=}\NormalTok{ SpotOptim(}
\NormalTok{    fun}\OperatorTok{=}\NormalTok{sphere,}
\NormalTok{    bounds}\OperatorTok{=}\NormalTok{[(}\OperatorTok{{-}}\DecValTok{5}\NormalTok{, }\DecValTok{5}\NormalTok{), (}\OperatorTok{{-}}\DecValTok{5}\NormalTok{, }\DecValTok{5}\NormalTok{)],}
\NormalTok{    max\_iter}\OperatorTok{=}\DecValTok{25}\NormalTok{,}
\NormalTok{    n\_initial}\OperatorTok{=}\DecValTok{10}\NormalTok{,}
\NormalTok{    seed}\OperatorTok{=}\DecValTok{0}\NormalTok{,}
\NormalTok{)}
\NormalTok{result }\OperatorTok{=}\NormalTok{ opt.optimize()}
\NormalTok{plot\_progress(opt, show}\OperatorTok{=}\VariableTok{False}\NormalTok{, figsize}\OperatorTok{=}\NormalTok{(}\DecValTok{6}\NormalTok{, }\DecValTok{4}\NormalTok{), log\_y}\OperatorTok{=}\VariableTok{True}\NormalTok{)}
\end{Highlighting}
\end{Shaded}

\begin{figure}[H]

\centering{

\pandocbounded{\includegraphics[keepaspectratio]{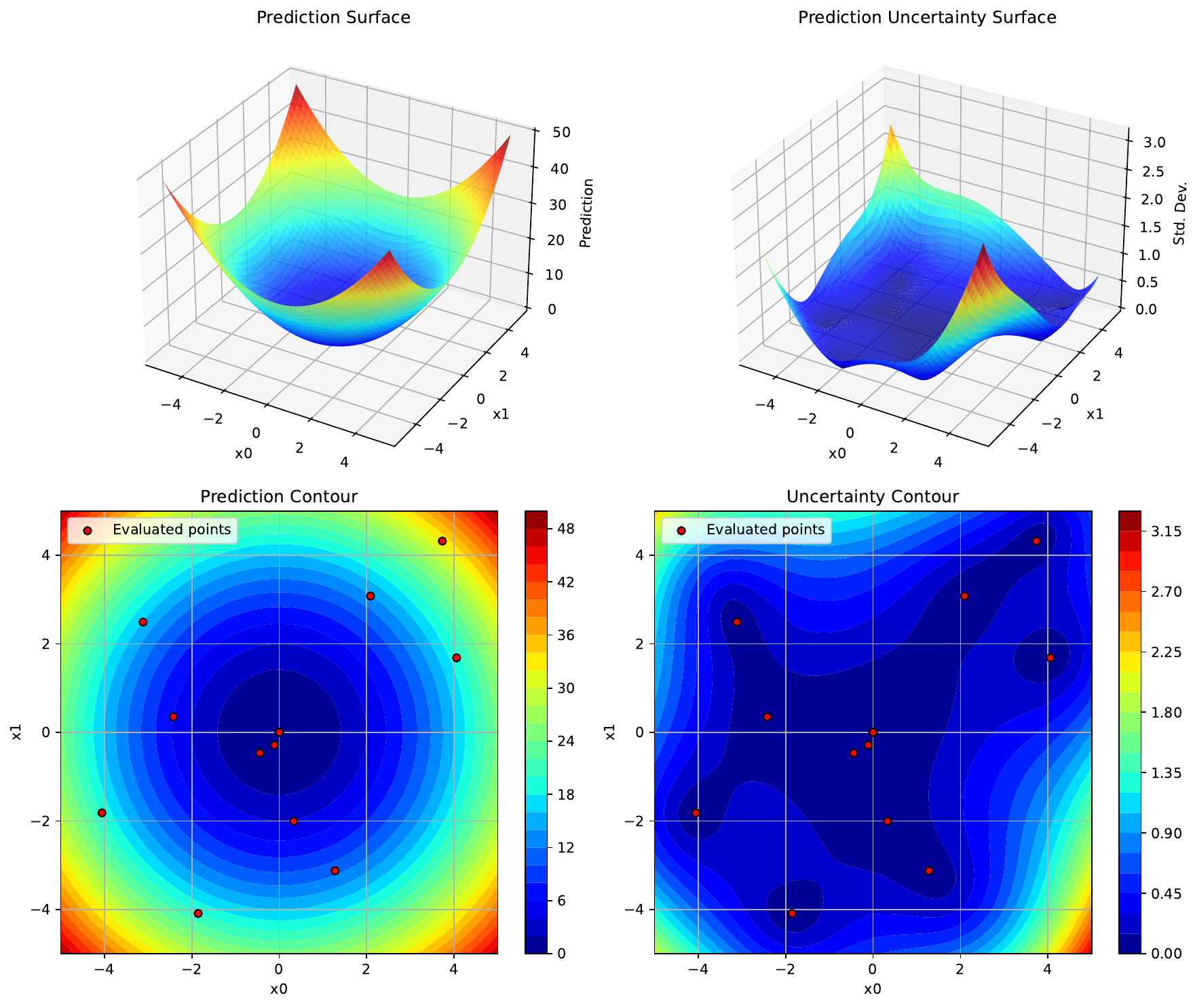}}

}

\caption{\label{fig-surrogate}Surrogate model for dimensions \(x_0\) and
\(x_1\). Top row: 3-D surfaces of predictions (left) and prediction
uncertainty (right). Bottom row: contour plots with evaluated points
overlaid.}

\end{figure}

Figure~\ref{fig-surrogate} shows the surrogate model fitted after
optimization. The top row displays 3-D surfaces of the predicted
objective value and the prediction uncertainty. The bottom row shows the
corresponding contour maps with the evaluated points overlaid as red
dots.

\begin{Shaded}
\begin{Highlighting}[]
\ImportTok{from}\NormalTok{ spotoptim.plot.visualization }\ImportTok{import}\NormalTok{ (}
\NormalTok{    plot\_surrogate}
\NormalTok{)}
\NormalTok{plot\_surrogate(opt, i}\OperatorTok{=}\DecValTok{0}\NormalTok{, j}\OperatorTok{=}\DecValTok{1}\NormalTok{, show}\OperatorTok{=}\VariableTok{False}\NormalTok{)}
\end{Highlighting}
\end{Shaded}

\begin{figure}[H]

\centering{

\pandocbounded{\includegraphics[keepaspectratio]{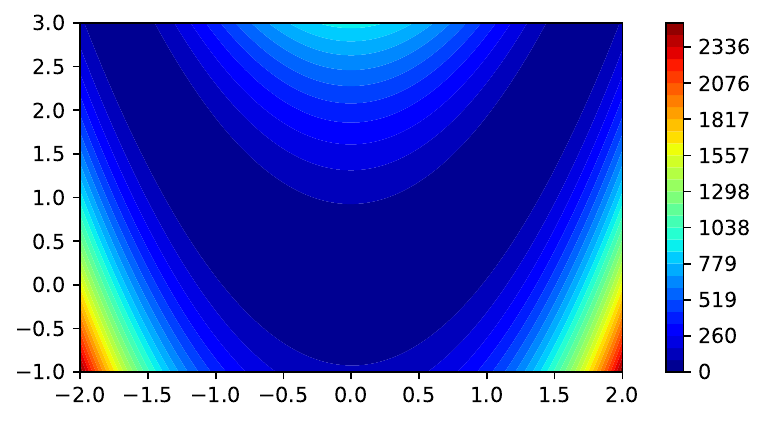}}

}

\caption{\label{fig-contour}Filled contour plot of the Rosenbrock
function over \([-2, 2] \times [-1, 3]\).}

\end{figure}

Figure~\ref{fig-contour} illustrates \texttt{simple\_contour} applied to
the Rosenbrock function. The function accepts any callable that maps a
\((1, 2)\) array to a scalar, making it convenient for quick inspection
of objective landscapes independently of an optimization run.

\begin{Shaded}
\begin{Highlighting}[]
\ImportTok{from}\NormalTok{ spotoptim.function }\ImportTok{import}\NormalTok{ rosenbrock}
\ImportTok{from}\NormalTok{ spotoptim.plot.contour }\ImportTok{import}\NormalTok{ (}
\NormalTok{    simple\_contour}
\NormalTok{)}
\NormalTok{simple\_contour(rosenbrock,}
\NormalTok{    min\_x}\OperatorTok{={-}}\DecValTok{2}\NormalTok{, max\_x}\OperatorTok{=}\DecValTok{2}\NormalTok{, min\_y}\OperatorTok{={-}}\DecValTok{1}\NormalTok{, max\_y}\OperatorTok{=}\DecValTok{3}\NormalTok{,}
\NormalTok{    n\_levels}\OperatorTok{=}\DecValTok{30}\NormalTok{)}
\end{Highlighting}
\end{Shaded}

\begin{figure}[H]

\centering{

\pandocbounded{\includegraphics[keepaspectratio]{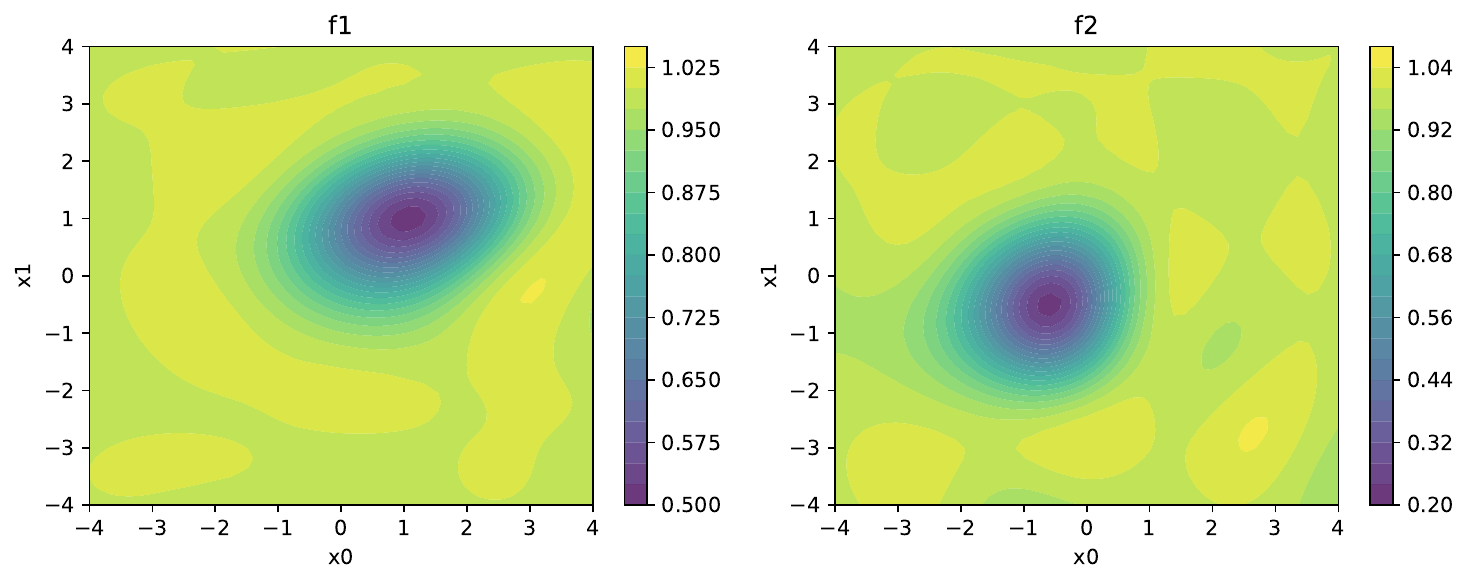}}

}

\caption{\label{fig-mo-contour}Surrogate contour plots for both
Fonseca--Fleming objectives, fitted from 50 random evaluations.}

\end{figure}

For multi-objective problems, \texttt{mo\_pareto\_optx\_plot} visualizes
Pareto-optimal points in the input space. The surrogate-based
visualization functions \texttt{mo\_xy\_contour} and
\texttt{mo\_xy\_surface} generate contour and surface plots for each
objective from fitted surrogate models. Figure~\ref{fig-mo-contour}
shows the contour view for two Kriging surrogates fitted to the
Fonseca--Fleming objectives.

\begin{Shaded}
\begin{Highlighting}[]
\ImportTok{import}\NormalTok{ numpy }\ImportTok{as}\NormalTok{ np}
\ImportTok{from}\NormalTok{ spotoptim.function }\ImportTok{import}\NormalTok{ fonseca\_fleming}
\ImportTok{from}\NormalTok{ spotoptim.surrogate }\ImportTok{import}\NormalTok{ Kriging}
\ImportTok{from}\NormalTok{ spotoptim.mo.pareto }\ImportTok{import}\NormalTok{ mo\_xy\_contour}

\NormalTok{rng }\OperatorTok{=}\NormalTok{ np.random.default\_rng(}\DecValTok{0}\NormalTok{)}
\NormalTok{X\_mo }\OperatorTok{=}\NormalTok{ rng.uniform(}\OperatorTok{{-}}\DecValTok{4}\NormalTok{, }\DecValTok{4}\NormalTok{, size}\OperatorTok{=}\NormalTok{(}\DecValTok{50}\NormalTok{, }\DecValTok{2}\NormalTok{))}
\NormalTok{Y\_mo }\OperatorTok{=}\NormalTok{ fonseca\_fleming(X\_mo)}

\NormalTok{m1 }\OperatorTok{=}\NormalTok{ Kriging()}
\NormalTok{m1.fit(X\_mo, Y\_mo[:, }\DecValTok{0}\NormalTok{])}
\NormalTok{m2 }\OperatorTok{=}\NormalTok{ Kriging()}
\NormalTok{m2.fit(X\_mo, Y\_mo[:, }\DecValTok{1}\NormalTok{])}
\NormalTok{mo\_xy\_contour(}
\NormalTok{    [m1, m2],}
\NormalTok{    bounds}\OperatorTok{=}\NormalTok{[(}\OperatorTok{{-}}\DecValTok{4}\NormalTok{, }\DecValTok{4}\NormalTok{), (}\OperatorTok{{-}}\DecValTok{4}\NormalTok{, }\DecValTok{4}\NormalTok{)],}
\NormalTok{    target\_names}\OperatorTok{=}\NormalTok{[}\StringTok{"f1"}\NormalTok{, }\StringTok{"f2"}\NormalTok{],}
\NormalTok{    feature\_names}\OperatorTok{=}\NormalTok{[}\StringTok{"x0"}\NormalTok{, }\StringTok{"x1"}\NormalTok{],}
\NormalTok{    resolution}\OperatorTok{=}\DecValTok{50}\NormalTok{,}
\NormalTok{)}
\end{Highlighting}
\end{Shaded}

\subsection{Utilities}\label{sec-utils}

The \texttt{utils} subpackage collects helper functions that support the
optimization loop and post-hoc analysis. \texttt{get\_boundaries}
computes column-wise minima and maxima, and
\texttt{map\_to\_original\_scale} transforms points from the \([0, 1]\)
unit hypercube back to the original variable ranges. PCA utilities
(\texttt{get\_pca}, \texttt{get\_pca\_topk}) perform PCA on evaluation
data and identify the features with the strongest loadings on the first
two components.

OCBA functions (\texttt{get\_ocba}, \texttt{get\_ranks}) implement the
OCBA algorithm for noisy optimization (Chen 1996; Chen et al. 2000;
Bartz-Beielstein and Friese 2011). Given current sample means,
variances, and an incremental budget, \texttt{get\_ocba} returns an
allocation vector that concentrates additional evaluations on the most
promising and most uncertain designs. \texttt{TorchStandardScaler}
standardizes PyTorch tensors to zero mean and unit variance, analogous
to scikit-learn's \texttt{StandardScaler}.

The TensorBoard integration (\texttt{utils/tensorboard.py}) provides
real-time monitoring of the optimization process. Setting
\texttt{tensorboard\_log=True} in the \texttt{SpotOptim} constructor
activates logging: at each iteration, the module writes scalar metrics
(current best objective value, last evaluation, success rate) and the
coordinates of the best design point to a TensorBoard event file. For
noisy optimization with repeated evaluations, additional statistics are
logged, including the best mean objective value and the variance at the
best design. Each evaluated hyperparameter configuration is also logged
via \texttt{add\_hparams}, which populates TensorBoard's HParams
dashboard and enables interactive comparison of configurations across
runs. The log directory defaults to
\texttt{runs/spotoptim\_YYYYMMDD\_HHMMSS} but can be customized via the
\texttt{tensorboard\_path} parameter. Setting
\texttt{tensorboard\_clean=True} removes all previous log directories
from the \texttt{runs} folder before a new optimization starts,
preventing clutter from accumulating across experiments. After
optimization completes, the writer is flushed and closed automatically.
The logs can then be viewed by running
\texttt{tensorboard\ -\/-logdir=runs} in a terminal and opening the
displayed URL in a browser. A minimal example that enables TensorBoard
logging\footnote{ View logs with:
  \texttt{tensorboard\ -\/-logdir=runs/my\_experiment}.} is:

\begin{Shaded}
\begin{Highlighting}[]
\ImportTok{from}\NormalTok{ spotoptim }\ImportTok{import}\NormalTok{ SpotOptim}
\ImportTok{from}\NormalTok{ spotoptim.function }\ImportTok{import}\NormalTok{ sphere}

\NormalTok{opt }\OperatorTok{=}\NormalTok{ SpotOptim(}
\NormalTok{    fun}\OperatorTok{=}\NormalTok{sphere,}
\NormalTok{    bounds}\OperatorTok{=}\NormalTok{[(}\OperatorTok{{-}}\DecValTok{5}\NormalTok{, }\DecValTok{5}\NormalTok{), (}\OperatorTok{{-}}\DecValTok{5}\NormalTok{, }\DecValTok{5}\NormalTok{)],}
\NormalTok{    max\_iter}\OperatorTok{=}\DecValTok{20}\NormalTok{,}
\NormalTok{    tensorboard\_log}\OperatorTok{=}\VariableTok{True}\NormalTok{,}
\NormalTok{    tensorboard\_clean}\OperatorTok{=}\VariableTok{True}\NormalTok{,}
\NormalTok{)}
\NormalTok{result }\OperatorTok{=}\NormalTok{ opt.optimize()}
\end{Highlighting}
\end{Shaded}

\subsection{Multi-Objective Optimization}\label{sec-mo}

The \texttt{mo} subpackage supports multi-objective optimization through
Pareto front analysis and scalarization. The
\texttt{is\_pareto\_efficient} function accepts a cost array of shape
\((n, m)\) and returns a boolean mask identifying the non-dominated
points. It works for any number of objectives and supports both
minimization and maximization.

Since the surrogate model operates on scalar objectives, multi-objective
functions must be scalarized before fitting. The \texttt{fun\_mo2so}
parameter of \texttt{SpotOptim} converts the \((n, m)\) output of the
objective function into a scalar \((n,)\) vector. The simplest
scalarization is a weighted sum:

\begin{Shaded}
\begin{Highlighting}[]
\ImportTok{import}\NormalTok{ numpy }\ImportTok{as}\NormalTok{ np}
\ImportTok{from}\NormalTok{ spotoptim }\ImportTok{import}\NormalTok{ SpotOptim}
\ImportTok{from}\NormalTok{ spotoptim.function }\ImportTok{import}\NormalTok{ (}
\NormalTok{    fonseca\_fleming}
\NormalTok{)}

\NormalTok{fun\_mo2so }\OperatorTok{=} \KeywordTok{lambda}\NormalTok{ y: np.}\BuiltInTok{sum}\NormalTok{(}
\NormalTok{    y }\OperatorTok{*}\NormalTok{ np.array([}\FloatTok{0.5}\NormalTok{, }\FloatTok{0.5}\NormalTok{]), axis}\OperatorTok{=}\DecValTok{1}
\NormalTok{)}

\NormalTok{opt }\OperatorTok{=}\NormalTok{ SpotOptim(}
\NormalTok{    fun}\OperatorTok{=}\NormalTok{fonseca\_fleming,}
\NormalTok{    bounds}\OperatorTok{=}\NormalTok{[(}\OperatorTok{{-}}\DecValTok{4}\NormalTok{, }\DecValTok{4}\NormalTok{), (}\OperatorTok{{-}}\DecValTok{4}\NormalTok{, }\DecValTok{4}\NormalTok{)],}
\NormalTok{    max\_iter}\OperatorTok{=}\DecValTok{30}\NormalTok{,}
\NormalTok{    n\_initial}\OperatorTok{=}\DecValTok{15}\NormalTok{,}
\NormalTok{    seed}\OperatorTok{=}\DecValTok{0}\NormalTok{,}
\NormalTok{    fun\_mo2so}\OperatorTok{=}\NormalTok{fun\_mo2so,}
\NormalTok{)}
\NormalTok{result }\OperatorTok{=}\NormalTok{ opt.optimize()}
\end{Highlighting}
\end{Shaded}

Different weight vectors trace different regions of the Pareto front.
For more sophisticated multi-objective handling, the
\texttt{spotdesirability} package provides desirability functions that
map multiple objectives onto a single composite scale while respecting
individual target values and importance weights (Bartz-Beielstein 2025a,
2025b).

\subsection{Hyperparameter Management}\label{sec-hyperparams}

The \texttt{hyperparameters} subpackage provides the
\texttt{ParameterSet} class, a fluent API for defining search spaces
with typed variables. Parameters are added through chained calls to
\texttt{add\_float()}, \texttt{add\_int()}, and \texttt{add\_factor()},
each specifying a name, bounds, default value, and optional
transformation.

\begin{Shaded}
\begin{Highlighting}[]
\ImportTok{from}\NormalTok{ spotoptim.hyperparameters }\OperatorTok{\textbackslash{}}
\NormalTok{    .parameters }\ImportTok{import}\NormalTok{ ParameterSet}

\NormalTok{ps }\OperatorTok{=}\NormalTok{ ParameterSet()}
\NormalTok{ps.add\_float(}
    \StringTok{"learning\_rate"}\NormalTok{,}
\NormalTok{    low}\OperatorTok{={-}}\DecValTok{5}\NormalTok{, high}\OperatorTok{={-}}\DecValTok{1}\NormalTok{, default}\OperatorTok{={-}}\DecValTok{3}\NormalTok{,}
\NormalTok{    transform}\OperatorTok{=}\StringTok{"log10"}\NormalTok{,}
\NormalTok{)}
\NormalTok{ps.add\_int(}
    \StringTok{"num\_layers"}\NormalTok{,}
\NormalTok{    low}\OperatorTok{=}\DecValTok{1}\NormalTok{, high}\OperatorTok{=}\DecValTok{5}\NormalTok{, default}\OperatorTok{=}\DecValTok{2}\NormalTok{,}
\NormalTok{)}
\NormalTok{ps.add\_float(}
    \StringTok{"dropout"}\NormalTok{,}
\NormalTok{    low}\OperatorTok{=}\FloatTok{0.0}\NormalTok{, high}\OperatorTok{=}\FloatTok{0.5}\NormalTok{, default}\OperatorTok{=}\FloatTok{0.1}\NormalTok{,}
\NormalTok{)}
\end{Highlighting}
\end{Shaded}

The properties \texttt{ps.bounds}, \texttt{ps.var\_type},
\texttt{ps.names()}, and \texttt{ps.var\_trans} map directly to the
corresponding \texttt{SpotOptim} constructor arguments, providing a
clean separation between search space definition and optimizer
configuration.

\subsection{Datasets}\label{sec-data}

The \texttt{data} subpackage provides PyTorch \texttt{Dataset} wrappers
for use in hyperparameter tuning workflows. \texttt{DiabetesDataset}
wraps the scikit-learn diabetes regression dataset (442 samples, 10
features) as a PyTorch \texttt{Dataset}, and
\texttt{get\_diabetes\_dataloaders()} creates train and test
\texttt{DataLoader} objects with configurable train/test split, batch
size, and optional feature scaling. These utilities simplify the setup
of neural network tuning experiments by providing ready-to-use data
pipelines.

\subsection{Model Inspection}\label{sec-inspection}

The \texttt{inspection} subpackage provides feature importance and
prediction diagnostics. \texttt{generate\_mdi()} trains a Random Forest
and extracts impurity-based feature importance scores.
\texttt{generate\_imp()} computes permutation importance by shuffling
each feature and measuring the degradation in model performance on a
held-out test set. \texttt{plot\_actual\_vs\_predicted()} creates
scatter plots comparing true values against model predictions, providing
a visual diagnostic of surrogate quality.

\begin{figure}[H]

\centering{

\pandocbounded{\includegraphics[keepaspectratio]{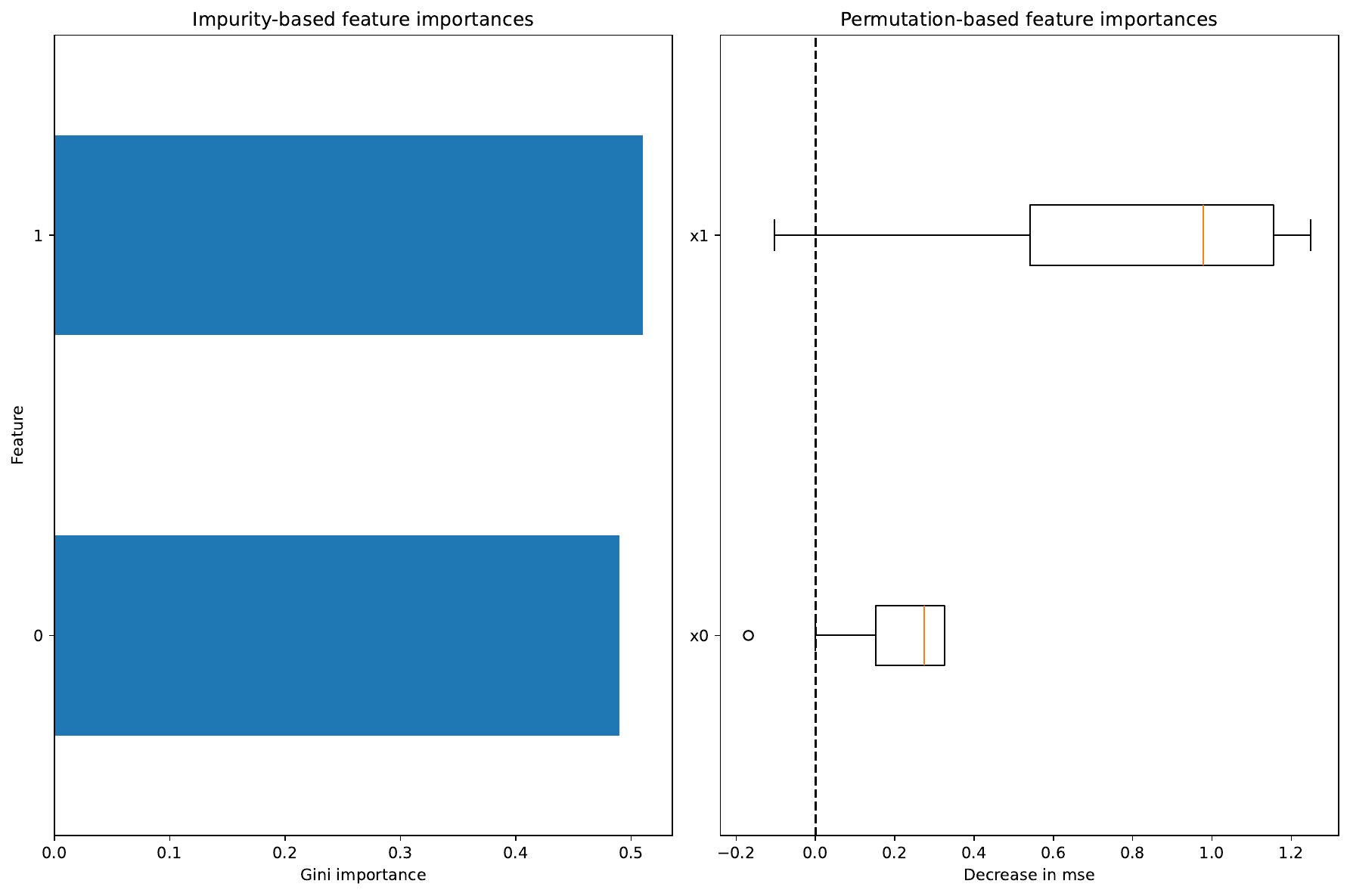}}

}

\caption{\label{fig-importances}Feature importances for the sphere
optimization. Left: impurity-based (Gini) importances from a Random
Forest. Right: permutation importances on the test set.}

\end{figure}

Figure~\ref{fig-importances} shows impurity-based (Gini) and
permutation-based feature importances for the sphere optimization from
Section~\ref{sec-plotting}. Both methods correctly identify \(x_0\) and
\(x_1\) as equally important, which is expected for the symmetric sphere
function.

\begin{Shaded}
\begin{Highlighting}[]
\ImportTok{from}\NormalTok{ sklearn.model\_selection }\ImportTok{import}\NormalTok{ (}
\NormalTok{    train\_test\_split}
\NormalTok{)}
\ImportTok{from}\NormalTok{ spotoptim.inspection }\ImportTok{import}\NormalTok{ (}
\NormalTok{    generate\_mdi, generate\_imp, plot\_importances}
\NormalTok{)}

\NormalTok{X\_tr, X\_te, y\_tr, y\_te }\OperatorTok{=}\NormalTok{ train\_test\_split(}
\NormalTok{    opt.X\_, opt.y\_, test\_size}\OperatorTok{=}\FloatTok{0.3}\NormalTok{, random\_state}\OperatorTok{=}\DecValTok{42}
\NormalTok{)}
\NormalTok{df\_mdi }\OperatorTok{=}\NormalTok{ generate\_mdi(X\_tr, y\_tr)}
\NormalTok{perm\_imp }\OperatorTok{=}\NormalTok{ generate\_imp(X\_tr, X\_te, y\_tr, y\_te)}
\NormalTok{plot\_importances(}
\NormalTok{    df\_mdi, perm\_imp, X\_te,}
\NormalTok{    feature\_names}\OperatorTok{=}\NormalTok{[}\StringTok{"x0"}\NormalTok{, }\StringTok{"x1"}\NormalTok{],}
\NormalTok{    show}\OperatorTok{=}\VariableTok{False}\NormalTok{,}
\NormalTok{)}
\end{Highlighting}
\end{Shaded}

\begin{figure}[H]

\centering{

\pandocbounded{\includegraphics[keepaspectratio]{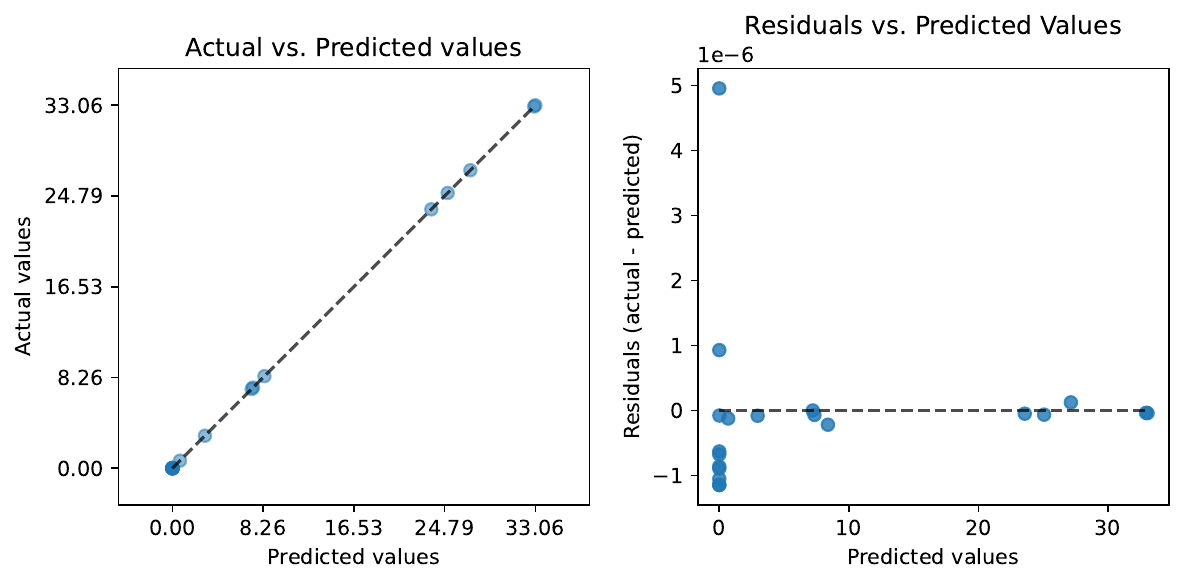}}

}

\caption{\label{fig-actual-vs-predicted}Surrogate prediction
diagnostics. Left: actual versus predicted objective values. Right:
residuals versus predicted values.}

\end{figure}

Figure~\ref{fig-actual-vs-predicted} compares the surrogate's
predictions against the true objective values for all evaluated points.
The left panel shows actual versus predicted values (points on the
diagonal indicate perfect agreement). The right panel shows residuals
versus predicted values.

\begin{Shaded}
\begin{Highlighting}[]
\ImportTok{from}\NormalTok{ spotoptim.inspection }\ImportTok{import}\NormalTok{ (}
\NormalTok{    plot\_actual\_vs\_predicted}
\NormalTok{)}

\NormalTok{y\_pred }\OperatorTok{=}\NormalTok{ opt.surrogate.predict(opt.X\_)}
\NormalTok{plot\_actual\_vs\_predicted(opt.y\_, y\_pred, show}\OperatorTok{=}\VariableTok{False}\NormalTok{)}
\end{Highlighting}
\end{Shaded}

\subsection{Factor Analysis}\label{sec-factor}

The \texttt{factor\_analyzer} subpackage provides tools for exploratory
factor analysis of high-dimensional optimization data. It is a port of
the \texttt{factor\_analyzer} package for Python\footnote{\url{https://factor-analyzer.readthedocs.io/en/latest/index.html}}.
Before running the analysis, suitability tests (\texttt{calculate\_kmo}
for the Kaiser-Meyer-Olkin measure,
\texttt{calculate\_bartlett\_sphericity} for Bartlett's test) check
whether the data has sufficient correlational structure. The
\texttt{FactorAnalyzer} class extracts latent factors with optional
varimax or promax rotation, helping to reveal the latent structure in
large parameter spaces.

\subsection{Exploratory Data Analysis}\label{sec-eda}

The \texttt{eda} subpackage provides quick visualization functions for
inspecting optimization data. \texttt{plot\_ip\_histograms()} creates a
grid of histograms for each variable, with categorical variables shown
as bar charts. Specific configurations (such as the best solution) can
be overlaid as vertical lines using the \texttt{add\_points} parameter.

\subsection{Triangulation Candidates}\label{sec-tricands}

The \texttt{tricands} module generates candidate points for acquisition
optimization by computing the Delaunay triangulation of existing
evaluated points (Gramacy et al. 2022). Interior candidates are placed
at simplex centroids, exploring gaps between existing evaluations.
Fringe candidates extend beyond the convex hull toward the search space
boundary, encouraging exploration of unexplored regions. The \texttt{p}
parameter controls the extension fraction, and \texttt{nmax} limits the
total number of candidates. This geometry-aware approach complements the
global search performed by differential evolution and is particularly
effective in low-to-moderate dimensions where the triangulation remains
computationally tractable.

\section{Hyperparameter Tuning with spotoptim}\label{sec-hpt}

A primary application of \texttt{spotoptim} is the tuning of machine
learning hyperparameters, where each function evaluation corresponds to
training and validating a model with a specific configuration. This
section demonstrates a complete neural network tuning workflow using the
diabetes regression dataset, an MLP architecture, and the
\texttt{spotoptim} optimization loop. To keep execution time manageable,
the number of training epochs and optimization iterations has been
reduced. In practice, longer training runs (50--200 epochs per
evaluation) and larger evaluation budgets (\texttt{max\_iter}
\(\geq 30\)) are necessary to obtain reliable results. The short
configuration used here is intended solely as a demonstration of the
workflow and API. The best hyperparameters found in such a short run
should not be considered representative.

The workflow follows five steps: define the search space, prepare the
dataset, define the objective function, run the optimization, and
analyze the results. This structure mirrors the hyperparameter tuning
methodology described in Bartz et al. (2022) and Bartz-Beielstein and
Zaefferer (2022), now implemented entirely in Python.

\subsection{Defining the Search Space}\label{defining-the-search-space}

The search space is defined using a \texttt{ParameterSet} that specifies
the hyperparameters to tune, their types, bounds, and transformations:

\begin{Shaded}
\begin{Highlighting}[]
\ImportTok{from}\NormalTok{ spotoptim.hyperparameters }\OperatorTok{\textbackslash{}}
\NormalTok{    .parameters }\ImportTok{import}\NormalTok{ ParameterSet}

\NormalTok{ps\_hpt }\OperatorTok{=}\NormalTok{ ParameterSet()}
\NormalTok{ps\_hpt.add\_float(}\StringTok{"lr"}\NormalTok{, low}\OperatorTok{=}\FloatTok{1e{-}5}\NormalTok{, high}\OperatorTok{=}\FloatTok{0.1}\NormalTok{,}
\NormalTok{    default}\OperatorTok{=}\FloatTok{0.001}\NormalTok{, transform}\OperatorTok{=}\StringTok{"log10"}\NormalTok{)}
\NormalTok{ps\_hpt.add\_int(}\StringTok{"l1"}\NormalTok{, low}\OperatorTok{=}\DecValTok{8}\NormalTok{, high}\OperatorTok{=}\DecValTok{128}\NormalTok{,}
\NormalTok{    default}\OperatorTok{=}\DecValTok{32}\NormalTok{)}
\NormalTok{ps\_hpt.add\_int(}\StringTok{"num\_hidden\_layers"}\NormalTok{,}
\NormalTok{    low}\OperatorTok{=}\DecValTok{1}\NormalTok{, high}\OperatorTok{=}\DecValTok{4}\NormalTok{, default}\OperatorTok{=}\DecValTok{2}\NormalTok{)}
\NormalTok{ps\_hpt.add\_float(}\StringTok{"dropout"}\NormalTok{, low}\OperatorTok{=}\FloatTok{0.0}\NormalTok{,}
\NormalTok{    high}\OperatorTok{=}\FloatTok{0.5}\NormalTok{, default}\OperatorTok{=}\FloatTok{0.1}\NormalTok{)}
\ControlFlowTok{for}\NormalTok{ n, t, b }\KeywordTok{in} \BuiltInTok{zip}\NormalTok{(}
\NormalTok{    ps\_hpt.names(),}
\NormalTok{    ps\_hpt.var\_type,}
\NormalTok{    ps\_hpt.bounds,}
\NormalTok{):}
    \BuiltInTok{print}\NormalTok{(}\SpecialStringTok{f"}\SpecialCharTok{\{}\NormalTok{n}\SpecialCharTok{\}}\SpecialStringTok{ (}\SpecialCharTok{\{}\NormalTok{t}\SpecialCharTok{\}}\SpecialStringTok{): }\SpecialCharTok{\{}\NormalTok{b}\SpecialCharTok{\}}\SpecialStringTok{"}\NormalTok{)}
\end{Highlighting}
\end{Shaded}

\begin{verbatim}
lr (float): (1e-05, 0.1)
l1 (int): (8, 128)
num_hidden_layers (int): (1, 4)
dropout (float): (0.0, 0.5)
\end{verbatim}

The learning rate bounds are specified in natural scale
(\([10^{-5}, 10^{-1}]\)); the \texttt{log10} transformation tells
SpotOptim to work internally in log space, so the surrogate models a
smooth landscape. SpotOptim automatically converts back to natural scale
before calling the objective function.

\subsection{Preparing the Dataset}\label{preparing-the-dataset}

The diabetes dataset is loaded and split into training and test sets
using the provided data loader utility:

\begin{Shaded}
\begin{Highlighting}[]
\ImportTok{from}\NormalTok{ spotoptim.data }\ImportTok{import}\NormalTok{ (}
\NormalTok{    get\_diabetes\_dataloaders,}
\NormalTok{)}

\NormalTok{train\_loader, test\_loader, scaler }\OperatorTok{=}\NormalTok{ (}
\NormalTok{    get\_diabetes\_dataloaders(}
\NormalTok{        test\_size}\OperatorTok{=}\FloatTok{0.2}\NormalTok{,}
\NormalTok{        batch\_size}\OperatorTok{=}\DecValTok{32}\NormalTok{,}
\NormalTok{        scale\_features}\OperatorTok{=}\VariableTok{True}\NormalTok{,}
\NormalTok{        random\_state}\OperatorTok{=}\DecValTok{0}\NormalTok{,}
\NormalTok{    )}
\NormalTok{)}
\BuiltInTok{print}\NormalTok{(}\SpecialStringTok{f"Training batches: }\SpecialCharTok{\{}\BuiltInTok{len}\NormalTok{(train\_loader)}\SpecialCharTok{\}}\SpecialStringTok{"}\NormalTok{)}
\BuiltInTok{print}\NormalTok{(}\SpecialStringTok{f"Test batches: }\SpecialCharTok{\{}\BuiltInTok{len}\NormalTok{(test\_loader)}\SpecialCharTok{\}}\SpecialStringTok{"}\NormalTok{)}
\end{Highlighting}
\end{Shaded}

\begin{verbatim}
Training batches: 12
Test batches: 3
\end{verbatim}

The \texttt{scale\_features=True} option standardizes input features to
zero mean and unit variance, which is important for neural network
training stability.

\subsection{Defining the Objective
Function}\label{defining-the-objective-function}

The objective function decodes hyperparameters from the search vector,
constructs a \texttt{LinearRegressor}, trains it on the training set,
and returns the mean squared error (MSE) on the test set. Because
SpotOptim applies the inverse of \texttt{var\_trans} before calling the
objective, the learning rate arrives in natural scale and can be used
directly. The number of epochs is set to 10 for this demo; production
runs should use 50--200.

\begin{Shaded}
\begin{Highlighting}[]
\ImportTok{import}\NormalTok{ numpy }\ImportTok{as}\NormalTok{ np}
\ImportTok{import}\NormalTok{ torch}
\ImportTok{from}\NormalTok{ spotoptim.nn }\ImportTok{import}\NormalTok{ LinearRegressor}

\NormalTok{N\_EPOCHS }\OperatorTok{=} \DecValTok{10}  \CommentTok{\# short demo; use 50{-}200}

\KeywordTok{def}\NormalTok{ nn\_objective(X):}
\NormalTok{    X }\OperatorTok{=}\NormalTok{ np.atleast\_2d(X)}
\NormalTok{    results }\OperatorTok{=}\NormalTok{ np.zeros(X.shape[}\DecValTok{0}\NormalTok{])}
    \ControlFlowTok{for}\NormalTok{ i }\KeywordTok{in} \BuiltInTok{range}\NormalTok{(X.shape[}\DecValTok{0}\NormalTok{]):}
\NormalTok{        lr }\OperatorTok{=}\NormalTok{ X[i, }\DecValTok{0}\NormalTok{]}
\NormalTok{        l1 }\OperatorTok{=} \BuiltInTok{int}\NormalTok{(X[i, }\DecValTok{1}\NormalTok{])}
\NormalTok{        n\_layers }\OperatorTok{=} \BuiltInTok{int}\NormalTok{(X[i, }\DecValTok{2}\NormalTok{])}
\NormalTok{        dropout }\OperatorTok{=}\NormalTok{ X[i, }\DecValTok{3}\NormalTok{]}
\NormalTok{        model }\OperatorTok{=}\NormalTok{ LinearRegressor(}
\NormalTok{            input\_dim}\OperatorTok{=}\DecValTok{10}\NormalTok{, output\_dim}\OperatorTok{=}\DecValTok{1}\NormalTok{,}
\NormalTok{            l1}\OperatorTok{=}\NormalTok{l1,}
\NormalTok{            num\_hidden\_layers}\OperatorTok{=}\NormalTok{n\_layers,}
\NormalTok{            activation}\OperatorTok{=}\StringTok{"ReLU"}\NormalTok{,}
\NormalTok{        )}
\NormalTok{        opt }\OperatorTok{=}\NormalTok{ torch.optim.Adam(}
\NormalTok{            model.parameters(), lr}\OperatorTok{=}\NormalTok{lr)}
\NormalTok{        loss\_fn }\OperatorTok{=}\NormalTok{ torch.nn.MSELoss()}
\NormalTok{        model.train()}
        \ControlFlowTok{for}\NormalTok{ epoch }\KeywordTok{in} \BuiltInTok{range}\NormalTok{(N\_EPOCHS):}
            \ControlFlowTok{for}\NormalTok{ xb, yb }\KeywordTok{in}\NormalTok{ train\_loader:}
\NormalTok{                opt.zero\_grad()}
\NormalTok{                loss }\OperatorTok{=}\NormalTok{ loss\_fn(model(xb), yb)}
\NormalTok{                loss.backward()}
\NormalTok{                opt.step()}
\NormalTok{        model.}\BuiltInTok{eval}\NormalTok{()}
\NormalTok{        total\_loss, n }\OperatorTok{=} \FloatTok{0.0}\NormalTok{, }\DecValTok{0}
        \ControlFlowTok{with}\NormalTok{ torch.no\_grad():}
            \ControlFlowTok{for}\NormalTok{ xb, yb }\KeywordTok{in}\NormalTok{ test\_loader:}
\NormalTok{                total\_loss }\OperatorTok{+=}\NormalTok{ (}
\NormalTok{                    loss\_fn(model(xb), yb)}
\NormalTok{                    .item() }\OperatorTok{*} \BuiltInTok{len}\NormalTok{(yb))}
\NormalTok{                n }\OperatorTok{+=} \BuiltInTok{len}\NormalTok{(yb)}
\NormalTok{        results[i] }\OperatorTok{=}\NormalTok{ total\_loss }\OperatorTok{/}\NormalTok{ n}
    \ControlFlowTok{return}\NormalTok{ results}
\end{Highlighting}
\end{Shaded}

The function follows \texttt{spotoptim}'s convention: it accepts a 2-D
array where each row is a configuration and returns a 1-D array of
objective values.

\subsection{Running the Optimization}\label{running-the-optimization}

With the search space and objective function defined, the optimization
is launched with a single call. The optimizer is configured with EI and
a small budget suitable for a demo. Production runs should increase
\texttt{max\_iter} to 30 or more.

\begin{Shaded}
\begin{Highlighting}[]
\ImportTok{from}\NormalTok{ spotoptim }\ImportTok{import}\NormalTok{ SpotOptim}

\NormalTok{opt\_hpt }\OperatorTok{=}\NormalTok{ SpotOptim(}
\NormalTok{    fun}\OperatorTok{=}\NormalTok{nn\_objective,}
\NormalTok{    bounds}\OperatorTok{=}\NormalTok{ps\_hpt.bounds,}
\NormalTok{    var\_type}\OperatorTok{=}\NormalTok{ps\_hpt.var\_type,}
\NormalTok{    var\_name}\OperatorTok{=}\NormalTok{ps\_hpt.names(),}
\NormalTok{    var\_trans}\OperatorTok{=}\NormalTok{ps\_hpt.var\_trans,}
\NormalTok{    acquisition}\OperatorTok{=}\StringTok{"ei"}\NormalTok{,}
\NormalTok{    max\_iter}\OperatorTok{=}\DecValTok{15}\NormalTok{, n\_initial}\OperatorTok{=}\DecValTok{8}\NormalTok{, seed}\OperatorTok{=}\DecValTok{0}\NormalTok{,}
\NormalTok{)}
\NormalTok{result\_hpt }\OperatorTok{=}\NormalTok{ opt\_hpt.optimize()}

\BuiltInTok{print}\NormalTok{(}\SpecialStringTok{f"Best MSE: }\SpecialCharTok{\{}\NormalTok{result\_hpt}\SpecialCharTok{.}\NormalTok{fun}\SpecialCharTok{:.4f\}}\SpecialStringTok{"}\NormalTok{)}
\BuiltInTok{print}\NormalTok{(}\SpecialStringTok{f"Evaluations: }\SpecialCharTok{\{}\NormalTok{result\_hpt}\SpecialCharTok{.}\NormalTok{nfev}\SpecialCharTok{\}}\SpecialStringTok{"}\NormalTok{)}
\BuiltInTok{print}\NormalTok{(}\StringTok{"Best config:"}\NormalTok{)}
\ControlFlowTok{for}\NormalTok{ n, v }\KeywordTok{in} \BuiltInTok{zip}\NormalTok{(}
\NormalTok{    ps\_hpt.names(), result\_hpt.x}
\NormalTok{):}
    \BuiltInTok{print}\NormalTok{(}\SpecialStringTok{f"  }\SpecialCharTok{\{}\NormalTok{n}\SpecialCharTok{\}}\SpecialStringTok{: }\SpecialCharTok{\{}\NormalTok{v}\SpecialCharTok{:.6g\}}\SpecialStringTok{"}\NormalTok{)}
\end{Highlighting}
\end{Shaded}

\begin{verbatim}
Best MSE: 3534.6855
Evaluations: 15
Best config:
  lr: 0.025114
  l1: 88
  num_hidden_layers: 2
  dropout: 0.408088
\end{verbatim}

The Kriging surrogate builds a model of the validation loss as a
function of the hyperparameters, and EI guides the search toward
configurations that are either predicted to perform well or that have
high uncertainty.

\subsection{Analyzing the Results}\label{analyzing-the-results}

\begin{figure}[H]

\centering{

\pandocbounded{\includegraphics[keepaspectratio]{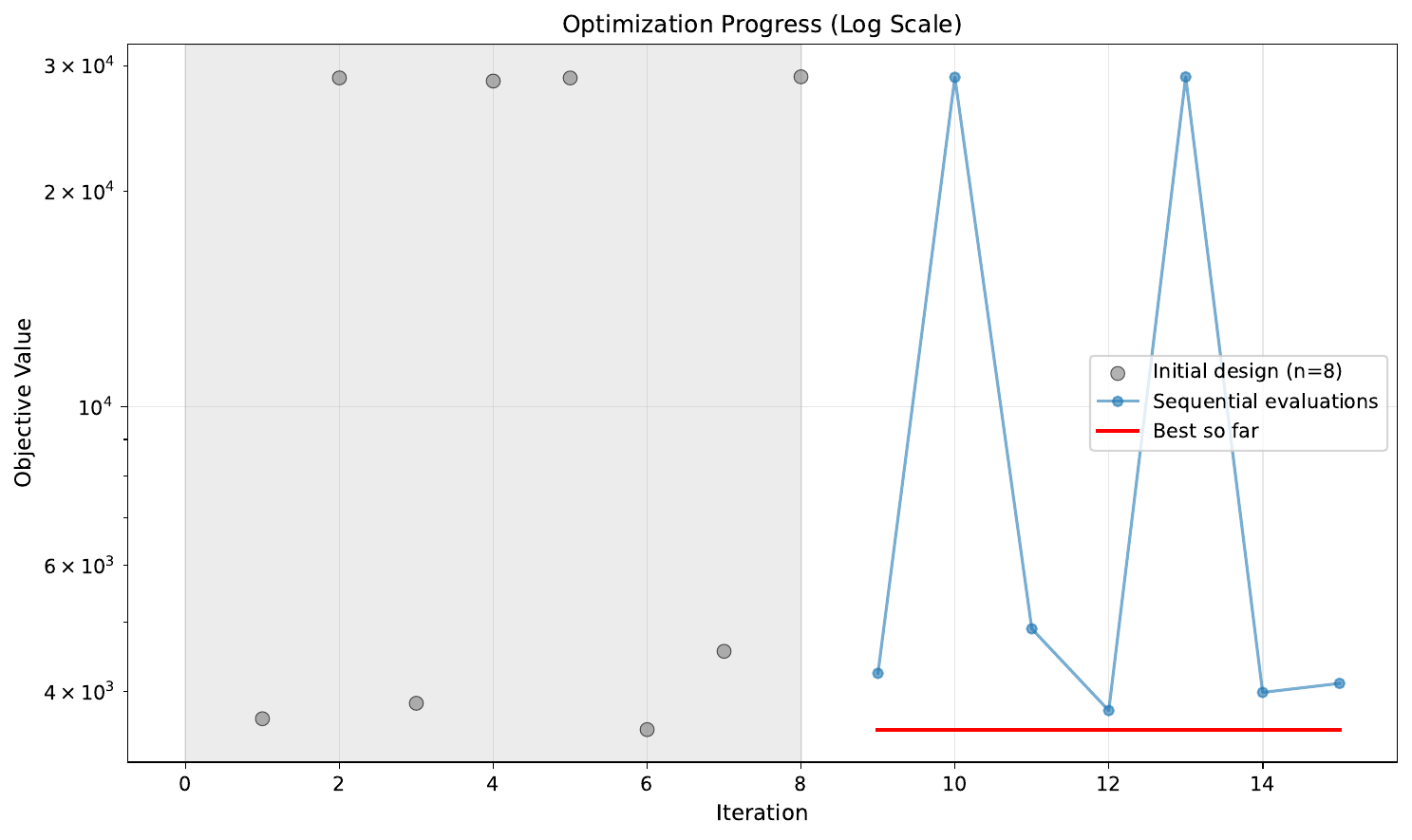}}

}

\caption{\label{fig-hpt-progress}Hyperparameter tuning progress (demo
run with reduced epochs and budget). The red curve shows the best
validation MSE found so far.}

\end{figure}

After optimization, the reporting utilities summarize which
hyperparameters were most influential and display the best
configuration. The progress plot (Figure~\ref{fig-hpt-progress}) shows
the convergence of the optimization.

\begin{Shaded}
\begin{Highlighting}[]
\ImportTok{from}\NormalTok{ spotoptim.plot.visualization }\ImportTok{import}\NormalTok{ (}
\NormalTok{    plot\_progress,}
\NormalTok{)}
\NormalTok{plot\_progress(opt\_hpt, show}\OperatorTok{=}\VariableTok{False}\NormalTok{, log\_y}\OperatorTok{=}\VariableTok{True}\NormalTok{)}
\end{Highlighting}
\end{Shaded}

\begin{figure}[H]

\centering{

\pandocbounded{\includegraphics[keepaspectratio]{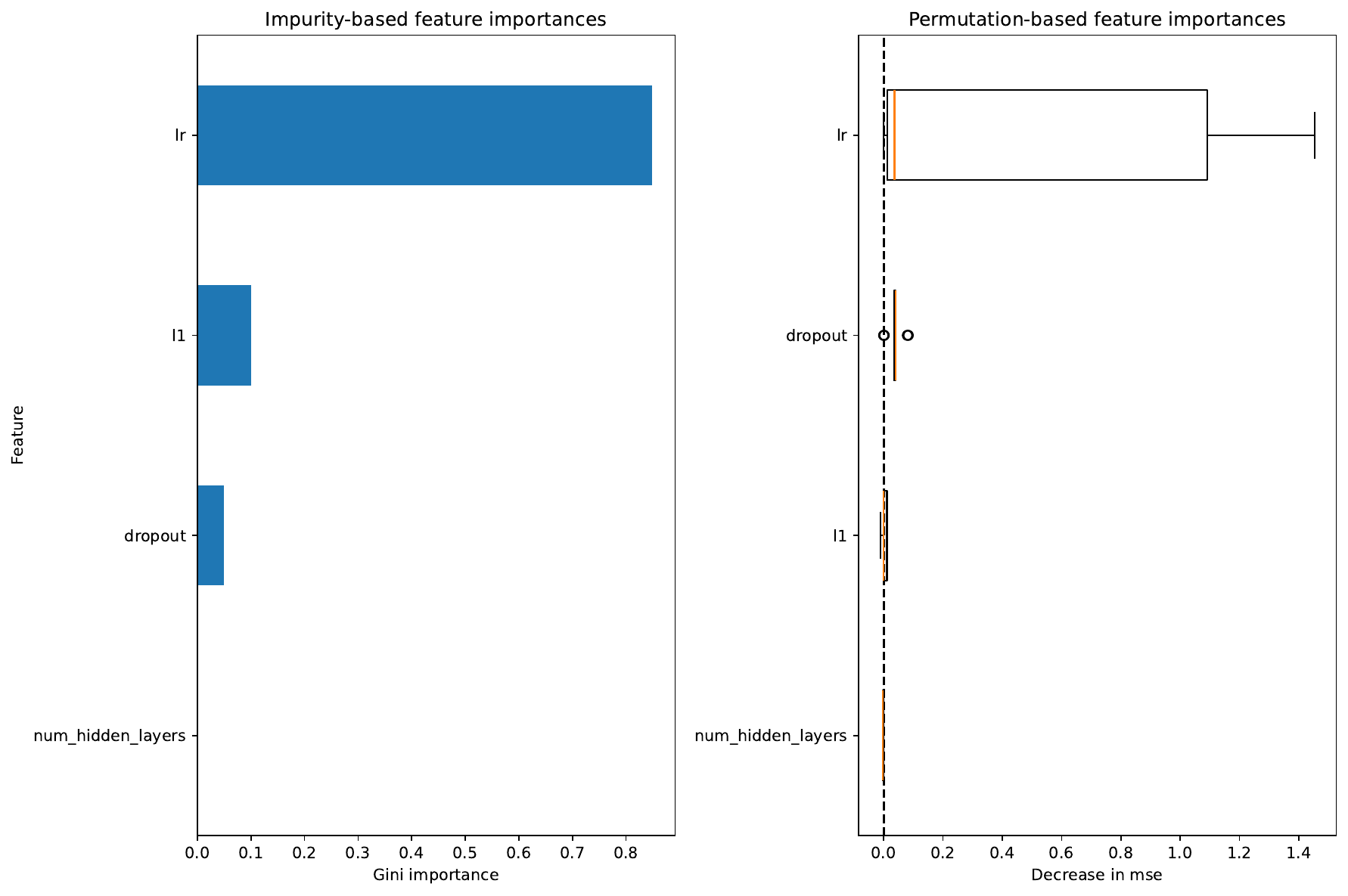}}

}

\caption{\label{fig-hpt-importances}Feature importances for the
hyperparameter tuning demo. Left: impurity-based (Gini) importances.
Right: permutation importances on the test set.}

\end{figure}

The feature importances (Figure~\ref{fig-hpt-importances}) reveal which
hyperparameters had the strongest influence on the validation loss.

\begin{Shaded}
\begin{Highlighting}[]
\ImportTok{from}\NormalTok{ sklearn.model\_selection }\ImportTok{import}\NormalTok{ (}
\NormalTok{    train\_test\_split,}
\NormalTok{)}
\ImportTok{from}\NormalTok{ spotoptim.inspection }\ImportTok{import}\NormalTok{ (}
\NormalTok{    generate\_mdi, generate\_imp,}
\NormalTok{    plot\_importances,}
\NormalTok{)}

\NormalTok{X\_tr, X\_te, y\_tr, y\_te }\OperatorTok{=}\NormalTok{ train\_test\_split(}
\NormalTok{    opt\_hpt.X\_, opt\_hpt.y\_,}
\NormalTok{    test\_size}\OperatorTok{=}\FloatTok{0.3}\NormalTok{, random\_state}\OperatorTok{=}\DecValTok{42}\NormalTok{,}
\NormalTok{)}
\NormalTok{df\_mdi }\OperatorTok{=}\NormalTok{ generate\_mdi(}
\NormalTok{    X\_tr, y\_tr,}
\NormalTok{    feature\_names}\OperatorTok{=}\NormalTok{ps\_hpt.names(),}
\NormalTok{)}
\NormalTok{perm\_imp }\OperatorTok{=}\NormalTok{ generate\_imp(}
\NormalTok{    X\_tr, X\_te, y\_tr, y\_te,}
\NormalTok{)}
\NormalTok{plot\_importances(}
\NormalTok{    df\_mdi, perm\_imp, X\_te,}
\NormalTok{    feature\_names}\OperatorTok{=}\NormalTok{ps\_hpt.names(),}
\NormalTok{    show}\OperatorTok{=}\VariableTok{False}\NormalTok{,}
\NormalTok{)}
\end{Highlighting}
\end{Shaded}

The \texttt{sensitivity\_spearman} function reports Spearman rank
correlations between each hyperparameter and the objective value, with
significance stars indicating statistical confidence. This helps the
practitioner understand which hyperparameters merit further
investigation and which can be fixed at their default values.

\begin{Shaded}
\begin{Highlighting}[]
\ImportTok{from}\NormalTok{ spotoptim.reporting.analysis }\ImportTok{import}\NormalTok{ (}
\NormalTok{    sensitivity\_spearman,}
\NormalTok{)}
\NormalTok{sensitivity\_spearman(opt\_hpt)}
\end{Highlighting}
\end{Shaded}

\begin{verbatim}

Sensitivity Analysis (Spearman Correlation):
--------------------------------------------------
  lr                  : -0.676 (p=0.006) **
  l1                  : -0.454 (p=0.089)
  num_hidden_layers   : +0.074 (p=0.794)
  dropout             : -0.533 (p=0.041) *
\end{verbatim}

The complete workflow described here can be compared with the
corresponding Ray Tune setup documented by Bartz-Beielstein (2023b).
While Ray Tune provides distributed scheduling across multiple machines,
\texttt{spotoptim} offers a more transparent, model-centric approach
where the user controls the surrogate model, acquisition function, and
experimental design. For single-machine workflows with moderate
evaluation budgets (tens to hundreds of configurations), the
surrogate-based approach is typically more sample-efficient than the
random or bandit-based strategies employed by Ray Tune's default
schedulers.

\section{Summary and Outlook}\label{sec-outlook}

This report has presented \texttt{spotoptim}, a Python package for
surrogate-model-based optimization of expensive black-box functions. The
package implements the SPO methodology with Kriging as the default
surrogate, EI and related acquisition functions, native support for
mixed variable types, noise-aware evaluation through repeated
evaluations and OCBA, and multi-objective extensions via Pareto analysis
and desirability functions. The architecture is designed around
scikit-learn compatibility for surrogates and scipy compatibility for
results, making the package interoperable with the broader Python
scientific computing ecosystem.

\texttt{spotoptim} represents the current generation of a two-decade
research lineage. It uses a modular architecture, structural typing
protocols, and comprehensive documentation. The package is part of an
ecosystem of related tools. For example, \texttt{spotdesirability}
provides desirability functions for multi-objective optimization,
enabling the user to express preferences over multiple objectives
through individual desirability curves and overall aggregation
(Bartz-Beielstein 2025a, 2025b), \texttt{spotforecast2} extends the
optimization framework to time-series forecasting, and
\texttt{spotforecast2\_safe} adds robustness guarantees for
safety-critical forecasting applications (Bartz-Beielstein and Bartz
2026). The current implementation runs strictly sequentially. The
emergence of free-threaded Python (PEP 703) opens the possibility of
true thread-level parallelism for objective evaluation, a natural
direction for future work on \texttt{spotoptim}. The \texttt{spotoptim}
package is open-source and available at
\url{https://gitlab.git.nrw/thk-f10/spotsevenlab/spotoptim} under the
AGPL-3.0 license. Documentation, including an API reference and a
comprehensive user guide with executable code examples, is hosted at
\url{https://advm1.gm.fh-koeln.de/~bartz/spotoptim/docs/index.html}.

\subsection*{Use of AI tools}\label{use-of-ai-tools}
\addcontentsline{toc}{subsection}{Use of AI tools}

Transparency notice: parts of this report were researched, drafted, and
translated with the support of artificial intelligence. This information
is provided voluntarily for complete transparency, as no legal
requirement mandates its disclosure.

\section*{References}\label{references}
\addcontentsline{toc}{section}{References}

\protect\phantomsection\label{refs}

\end{document}